\documentclass[10pt,twocolumn]{article}

\usepackage{cite}
\usepackage{amsmath,amssymb,amsfonts}
\usepackage{algorithmic}
\usepackage{graphicx}
\usepackage{array}
\usepackage{caption}
\usepackage{longtable}
\usepackage{placeins}
\usepackage{url}
\usepackage{textcomp}
\usepackage{lipsum}
\usepackage{xcolor}
\usepackage{algorithm}
\usepackage{tabularx}
\usepackage{float}
\usepackage{graphicx} 
\usepackage{tabularx}
\usepackage{makecell}

\usepackage[normalem]{ulem}

\title{TWSSenti: A Novel Hybrid Framework for Topic-Wise Sentiment Analysis on Social Media Using Transformer Models}

\author{
\textbf{Aish Albladi}$^{1}$,
\textbf{Md Kaosar Uddin}$^{2}$,
\textbf{Minarul Islam}$^{2}$,
\textbf{Cheryl Seals}$^{2}$\\[0.5em]
$^{1}$ Jouf University, Sakakah, Aljouf 2014, Saudi Arabia\\
$^{2}$Auburn University, AL, USA\\
\texttt{aish.cs@ju.edu.sa}
}
\date{}
\begin{document}

\twocolumn[
\maketitle
\begin{abstract}
Sentiment analysis is a crucial task in natural language processing (NLP) that enables the extraction of meaningful insights from textual data, particularly from dynamic platforms like Twitter and IMDB. This study explores a hybrid framework combining transformer-based models, specifically BERT, GPT-2, RoBERTa, XLNet, and DistilBERT, to improve sentiment classification accuracy and robustness. The framework addresses challenges such as noisy data, contextual ambiguity, and generalization across diverse datasets by leveraging the unique strengths of these models. BERT captures bidirectional context, GPT-2 enhances generative capabilities, RoBERTa optimizes contextual understanding with larger corpora and dynamic masking, XLNet models dependency through permutation-based learning, and DistilBERT offers efficiency with reduced computational overhead while maintaining high accuracy. We demonstrate text cleaning, tokenization, and feature extraction using Term Frequency Inverse Document Frequency (TF-IDF) and Bag of Words (BoW), ensuring high-quality input data for the models. The hybrid approach was evaluated on benchmark datasets Sentiment140 and IMDB, achieving superior accuracy rates of 94\% and 95\%, respectively, outperforming standalone models. Experimental results show that TWSSenti achieves accuracies of 94\% on Sentiment140 and 95\% on IMDB. This corresponds to a relative improvement of 3.2\% and 2.7\%, respectively, over strong transformer baselines such as BERT and XLNet, and more than a 10\% gain compared with traditional machine learning models. These improvements demonstrate the effectiveness of the proposed hybrid fusion approach for fine-grained sentiment analysis. The results validate the effectiveness of combining multiple transformer models in ensemble-like setups to address the limitations of individual architectures. This research highlights its applicability to real-world tasks such as social media monitoring, customer sentiment analysis, and public opinion tracking which offer a pathway for future advancements in hybrid NLP frameworks.
\end{abstract}
\vspace{0.5em}
]


\section{Introduction}
In the age of digital transformation, social media platforms like Twitter and review sites such as IMDB have become essential spaces where individuals express their opinions, emotions, and sentiments on a diverse range of topics \cite{sharma2024review}. The exponential growth of user-generated content has created an immense repository of unstructured textual data which offers unprecedented insights into consumer behavior, public sentiment, and broader societal trends \cite{chang2014understanding, hu2014toward}. This wealth of information is invaluable for businesses, policymakers, and researchers, as it enables them to make data-driven decisions, monitor brand perception, enhance customer satisfaction, and gain a deeper understanding of social trends in real-time \cite{perera2021big, zong2024ai}. Sentiment analysis, also referred to as opinion mining and a crucial subset of natural language processing (NLP), has emerged as a transformative tool for harnessing the potential of this data \cite{jim2024recent, anderson2024analyzing}. By extracting and categorizing the emotions and opinions embedded in textual data, sentiment analysis allows for the systematic interpretation of attitudes and emotional tones within large volumes of text \cite{wankhade2022survey, uddin2026divergence}.

At its core, sentiment analysis aims to evaluate text polarity—whether the expressed sentiment is positive, negative, or neutral—while also identifying more nuanced emotional tones such as joy, anger, or sarcasm \cite{al2024systematic}. This makes it an indispensable tool for a variety of applications, including customer feedback analysis, brand monitoring, and public opinion assessment. For instance, organizations can use sentiment analysis to track consumer satisfaction by analyzing product reviews, and public response to marketing campaigns, or monitor brand reputation during significant events \cite{zimbra2018state}. Similarly, policymakers can leverage sentiment analysis to understand public opinion on pressing social and political issues that reflects public sentiment \cite{ashayeri2024unraveling}. Beyond its practical applications, sentiment analysis plays a critical role in transforming unstructured data into structured, actionable insights \cite{williams2023natural}. However, analyzing text from sources such as social media presents inherent challenges due to its informal nature, including the use of slang, abbreviations, emojis, and cultural variations in sentiment expression \cite{alfreihat2024emo, sundaram2023systematic}. These complexities require advanced methodologies capable of handling noisy, informal, and often ambiguous text. We address these challenges by proposing a hybrid framework that combines state-of-the-art transformer models—BERT, GPT-2, RoBERTa, XLNet, and DistilBERT—to leverage their complementary strengths for robust sentiment analysis. Our approach integrates advanced preprocessing techniques to handle noisy and informal text. This comprehensive methodology enhances sentiment classification accuracy, reduces misclassifications, and ensures scalability across diverse datasets which makes it adaptable for real-world applications such as social media monitoring, customer feedback analysis, and public sentiment tracking.

Traditional sentiment analysis approaches, including machine learning algorithms like Support Vector Machines (SVM) and Naïve Bayes, as well as lexicon-based methods, have been widely employed for text classification \cite{srivastava2022comparative, naithani2023realization,uddin2025alpha, sham2022climate}. However, these methods often struggle to handle the complexity of natural language, such as contextual nuances, ambiguous expressions, and informal language commonly used on platforms like Twitter \cite{chaudhary2023review, jim2024recent,uddin2020comparative}. With the advent of deep learning and, more recently, transformer-based models, the landscape of sentiment analysis has undergone a significant transformation. Despite their success, these models are not without limitations. Challenges such as overfitting, computational inefficiency, and poor generalization across datasets persist, particularly when applied to real-world, large-scale datasets like Sentiment140 and IMDB. The motivation behind this research stems from the need to address these limitations and further enhance the capabilities of transformer-based models for sentiment analysis. Individual transformer models, while powerful, often fail to capture all the intricacies of textual data due to their reliance on specific training paradigms. Additionally, standalone models are prone to overfitting when trained on smaller or imbalanced datasets that lead to suboptimal performance in real-world applications.

Despite significant advancements in transformer-based models, the growing diversity and volume of user-generated content present ongoing challenges for sentiment analysis. Real-world applications, such as social media monitoring and customer opinion tracking, require models that not only achieve high accuracy but also scale efficiently across large, dynamic datasets. Texts from different domains, such as reviews, social media posts, and opinion articles, vary significantly in structure, length, and linguistic style, adding complexity to the analysis. For instance, while short-form texts like tweets often include abbreviations, emojis, and slang, long-form reviews demand an understanding of nuanced and context-dependent sentiments. Existing standalone models, though powerful, often fail to generalize effectively across these varying text types. To overcome these limitations, this study introduces a robust hybrid framework that combines transformer-based models with advanced preprocessing techniques, enabling scalability, adaptability, and improved performance across diverse real-world datasets. By addressing these challenges, the proposed approach not only enhances sentiment classification accuracy but also ensures the practical applicability of sentiment analysis in domains like marketing analytics, public opinion tracking, and social media research.

While sentiment analysis using transformer models has been widely explored, our hybrid framework introduces a novel ensemble fusion strategy that balances accuracy, computational efficiency, and domain adaptability. Unlike conventional single-model approaches, our method leverages complementary strengths of multiple transformers, including BERT, GPT-2, RoBERTa, XLNet, and DistilBERT. The weighted ensemble strategy ensures that models contributing higher predictive confidence influence the final classification more significantly, improving overall sentiment detection. To further distinguish our work, we provide a comparative analysis with existing ensemble-based sentiment models and emphasize how our hybrid approach achieves superior performance with optimized resource consumption. These distinctions underscore the novelty and contributions of our framework within the field of sentiment analysis.

To the best of our knowledge, the aforementioned research articles have overlooked the combination of hybrid models for sentiment analysis on Twitter data. To bridge these gaps, this study proposes a novel hybrid model that combines the strengths of BERT, GPT-2, RoBERTa, XLNet, and DistilBERT, respectively. By integrating these components, the proposed approach aims to achieve superior accuracy and robustness in sentiment classification tasks. This ensemble strategy not only addresses the limitations of individual models but also provides a scalable solution for large-scale sentiment analysis. The research evaluates the hybrid model's performance on two benchmark datasets, Sentiment140 and IMDB, representing social media and movie review data, respectively. Comprehensive experiments demonstrate that the hybrid model significantly outperforms individual transformer models, achieving accuracy rates of 94\% on Sentiment140 and 95\% on IMDB. The study also investigates the impact of preprocessing techniques, such as tokenization, stemming, and noise removal, as well as feature extraction methods like TF-IDF and Bag of Words, on overall model performance. The results underscore the importance of a carefully designed pipeline in achieving high accuracy and robustness in sentiment analysis.

\begin{figure}[htbp]
 \centering
  \includegraphics[width= 0.8 \linewidth]{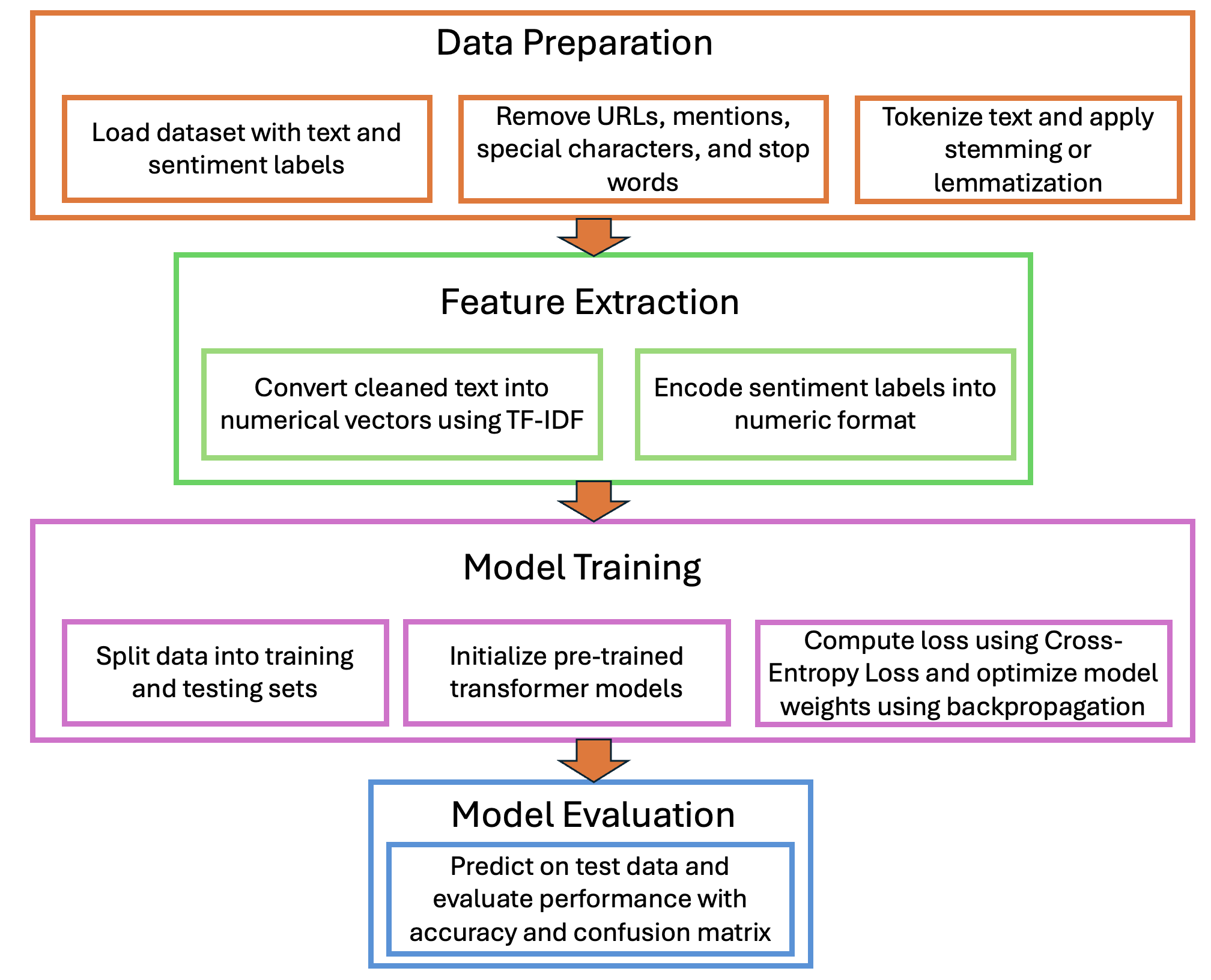} 
  \caption{TWSSenti overall design.} 
  \label{fig:overall}
\end{figure}

We demonstrate the overall workflow for sentiment analysis using transformer-based models as shown in Fig. \ref{fig:overall}. The process begins with data preparation, where the dataset containing text and sentiment labels is loaded. This stage involves cleaning the text by removing URLs, mentions, special characters, and stop words, followed by tokenizing the text and applying stemming or lemmatization to standardize the data. Next, in the feature Eetraction stage, the cleaned text is converted into numerical representations using techniques like TF-IDF, while sentiment labels are encoded into numeric formats for compatibility with machine learning models. The model training phase involves splitting the data into training and testing sets, initializing pre-trained transformer models, and training them by computing loss with cross-entropy loss and optimizing model weights using backpropagation. Finally, the model evaluation phase evaluates the model's performance on the test data, predicting sentiment labels and calculating metrics like accuracy and confusion matrix to assess the model's effectiveness in sentiment classification. This streamlined workflow ensures efficient and accurate sentiment analysis.

\textbf{Key Contributions:} The key contributions of this research are summarized as follows:
\begin{enumerate}

    \item \textbf{Impact of Preprocessing and Feature Engineering:} We propose a systematic evaluation of preprocessing techniques (e.g., tokenization, stemming, and noise removal) and feature extraction methods (TF-IDF and Bag of Words) on model effectiveness.
    
    \item \textbf{Proposed Hybrid Model:} We demonstrate a hybrid model combining BERT, XLNet, and CatBoost to address the limitations of individual transformer models and enhance sentiment classification accuracy and generalization.
    
    \item \textbf{Comprehensive Performance Evaluation:} We compare the performance of hybrid model against state-of-the-art transformer-based models (BERT, GPT-2, RoBERTa, XLNet, DistilBERT) on Sentiment140 and IMDB datasets which highlights its superior performance.
    
    \item \textbf{Insights for Real-World Applications:} We validate the hybrid model's applicability in real-time sentiment analysis scenarios, such as social media monitoring and consumer insights, providing practical solutions for businesses and researchers.

    \item \textbf{Recommendations for Future Research:} We discuss the challenges, including handling sarcasm, imbalanced datasets, and cross-domain adaptability, along with suggestions for improving hybrid approaches and advancing the field of sentiment analysis.
\end{enumerate} 

The rest of the paper is organized as follows: In Section \ref{related_work:graphical}, the literature review is discussed. Section \ref{methodology:graphical} demonstrates the Methodology; Section \ref{discussion:graphical} discuss the challenges and future research directions regarding NLPs for sentiment analysis on Twitter data, Section \ref{conclusion:graphical} finally concludes the paper.

\section{Literature review}\label{related_work:graphical}
Sentiment analysis has witnessed significant evolution over the years, transitioning from traditional machine learning techniques to advanced deep learning and transformer-based architectures. While these advancements have significantly improved sentiment classification accuracy, challenges remain in handling context-dependent sentiment, sarcasm, and imbalanced datasets. This section reviews the progression of sentiment analysis techniques, including traditional machine learning approaches, deep learning methods, transformer-based models, and hybrid frameworks, with their contributions, limitations, and gaps in the field.

\subsection{Traditional Machine Learning Approaches}
Traditional machine learning techniques were among the earliest methods employed for sentiment analysis. Algorithms like Support Vector Machines (SVM), Naive Bayes, and decision trees were widely used for text classification tasks \cite{li2022review, zhang2014machine, shayaa2018sentiment, uddin2025spd}. These approaches relied heavily on handcrafted features, such as n-grams, Bag of Words (BoW), and Term Frequency-Inverse Document Frequency (TF-IDF), to represent textual data numerically \cite{lukwaro2024review, mahadevkar2024exploring}. For instance, BoW represented text as a sparse vector of word occurrences, while TF-IDF weighted words based on their importance across a document corpus \cite{lin2024deep}. Although these methods provided foundational insights into text classification, they were inherently limited by their inability to capture semantic relationships and contextual dependencies in text. For example, while SVMs and Naive Bayes achieved reasonable accuracy, their performance declined significantly on larger or more complex datasets. The inability to handle long-range dependencies in text further constrained their scalability. Moreover, these methods required extensive feature engineering, making them labor-intensive and less adaptive to changes in data or domain. The authors \cite{barbierato2024challenges, raines2024enhancing} highlighted that traditional machine learning approaches often struggled to generalize well across diverse datasets, thereby limiting their applicability to real-world sentiment analysis tasks. These limitations underscored the need for more advanced, context-aware models capable of learning features automatically.

\subsection{Deep Learning Models}
Deep learning marked a significant advancement in sentiment analysis, surpassing traditional machine learning methods that relied heavily on handcrafted features such as Bag of Words (BoW), TF-IDF, and n-grams. Models like Long Short-Term Memory (LSTM) networks, Bidirectional LSTMs (BiLSTMs), and Convolutional Neural Networks (CNNs) enabled the automatic extraction of hierarchical text representations, reducing dependence on manual feature engineering \cite{abimbola2024enhancing}. LSTMs introduced memory cells to capture long-term dependencies, making them effective for sequential data like text \cite{devika2024book}, while BiLSTMs further improved performance by considering both past and future contexts \cite{wankhade2024cbmafm}. Researchers demonstrated the benefits of combining BiLSTMs with pre-trained embeddings like Word2Vec and GloVe, achieving enhanced classification accuracy \cite{shuqin2024deep, jiang2024dcasam, manasa2024detection}. However, LSTMs and BiLSTMs remained computationally expensive and struggled with vanishing gradients in very long sequences. CNNs excelled at capturing local patterns such as phrases and n-grams but lacked the ability to model sequential or bidirectional contextual relationships effectively \cite{dey2024machine}. Despite these advancements, deep learning models faced challenges in detecting sarcasm, mixed emotions, and handling highly imbalanced datasets \cite{boutsikaris2024comparative}. These limitations led to the development of transformer-based models, which introduced self-attention mechanisms to capture global word relationships, addressing the shortcomings of earlier approaches.

\subsection{Transformer-Based Models}
The advent of transformer-based architectures revolutionized sentiment analysis by introducing mechanisms that captured long-range dependencies and bidirectional context. Models like BERT, GPT-2, RoBERTa, XLNet, and DistilBERT set new benchmarks in text classification tasks \cite{raiaan2024review, sufi2024sustainable}. BERT, for instance, utilized a bidirectional transformer to capture context from both directions, enabling a deeper understanding of sentence structure and semantics \cite{zhou2024comprehensive}. RoBERTa improved upon BERT by training on larger datasets and employing dynamic masking techniques, resulting in enhanced performance on sentiment analysis tasks \cite{jin2024wordtransabsa}. GPT-2, with its autoregressive structure, excelled in generative tasks but was less effective for classification tasks due to its unidirectional context modeling \cite{oralbekova2023contemporary,uddin2025spd,uddin2026divergence}. XLNet introduced permutation-based modeling, which enhanced bidirectional learning by considering all possible word orderings in a sentence \cite{aljrees2024contradiction,uddin2025alpha}. This study made XLNet particularly robust for long-form text, such as IMDB movie reviews, where understanding global context was critical. The authors  demonstrated that transformer-based models like BERT and RoBERTa consistently outperformed traditional and deep learning models, achieving state-of-the-art results on datasets like Sentiment140 and IMDB \cite{tan2023roberta}. However, transformer models are not without limitations. These architectures are computationally intensive, requiring significant resources for training and fine-tuning. They are also prone to overfitting on small or imbalanced datasets, where a lack of data diversity can limit generalization. Moreover, individual transformer models often struggle with domain-specific challenges, such as sarcasm, slang, or abbreviations common in social media text. While their self-attention mechanisms allow for expressive encoding of relationships within text, they lack adaptability to diverse linguistic and cultural contexts without domain-specific fine-tuning.

\subsection{Hybrid Models}
To address the limitations of standalone models, recent research has explored hybrid approaches that combine the strengths of multiple methods. Hybrid models integrate transformer-based architectures with other techniques to enhance performance and robustness. For instance, the researchers proposed a hybrid framework that combined BERT with lexicon-based methods for Twitter sentiment analysis \cite{catelli2022lexicon}. However, while hybrid models often outperform standalone methods, they introduce additional complexity in model integration and are computationally demanding. Existing hybrid approaches also face challenges in optimizing computational efficiency and addressing overfitting, particularly when applied to large-scale datasets. Additionally, the integration of multiple models often requires extensive fine-tuning, which can limit scalability. Despite these challenges, hybrid models represent a promising direction for sentiment analysis, as they effectively combine the strengths of multiple architectures to address context-dependent sentiment, sarcasm, and imbalanced data.

In this paper, we address the aforementioned limitations observed in previous works on sentiment analysis by proposing a robust hybrid framework that leverages multiple transformer-based models, including BERT, GPT-2, RoBERTa, XLNet, and DistilBERT. Unlike traditional machine learning approaches and standalone deep learning models, our framework effectively captures bidirectional context, long-range dependencies, and nuanced sentiments, such as sarcasm and mixed emotions, across diverse datasets. Additionally, we mitigate challenges like overfitting on small or imbalanced datasets and improve generalization through advanced preprocessing and fine-tuning techniques. While prior studies have either focused on individual models or simplistic hybrid approaches, our work integrates the complementary strengths of multiple transformers to deliver superior performance on noisy, context-dependent, and domain-specific text data, demonstrating scalability and adaptability in real-world applications.

\section{Methodology}\label{methodology:graphical} 
The methodology employed in this study focuses on addressing the challenges of sentiment analysis by leveraging advanced transformer models. This section comprehensively details the data collection, preprocessing, feature extraction, hybrid model development, training process, evaluation strategies, and real-world applications. The proposed methodology ensures robust sentiment classification by overcoming the limitations of individual models, such as overfitting, limited contextual understanding, and lack of generalization.

\subsection{Dataset Description}
To ensure a comprehensive evaluation of the proposed hybrid framework, two diverse and widely recognized datasets were selected: Sentiment140 and IMDB Movie Reviews \cite{tan2023survey}. These datasets were chosen to test the model’s performance across different text types, domains, and complexities, enabling a balanced assessment of its scalability and generalizability.

The Sentiment140 dataset consists of 1.6 million labeled tweets, categorized as positive, negative, or neutral. Tweets in this dataset reflect the informal nature of social media communication, characterized by abbreviations, slang, emojis, hashtags, and frequent misspellings. This dataset presents unique challenges for sentiment analysis, as the text is often noisy, short, and context-dependent. For example, tweets frequently contain sarcasm or informal expressions that are difficult to capture using conventional methods. Sentiment140 serves as a robust platform for evaluating the model’s ability to handle noisy, unstructured, and abbreviated text, which is crucial for real-world applications like social media monitoring and customer feedback analysis. In contrast, the IMDB Movie Reviews dataset consists of 50,000 labeled movie reviews, split evenly into positive and negative sentiments. Unlike Sentiment140, these reviews are longer, more structured, and written in a formal style. They often include nuanced expressions of sentiment, with complex sentence structures and context-dependent sentiment shifts. For instance, a single review might begin with a negative critique and transition to an overall positive sentiment by the end. This dataset is particularly valuable for evaluating the model’s capacity to process and analyze long-form text, where understanding contextual relationships across multiple sentences is essential. Additionally, IMDB reviews often contain domain-specific terminology related to films, actors, and genres, further challenging the model to adapt to different content types.

By using these two datasets, the study evaluates the model’s performance across both short and long text formats, ensuring its applicability across a range of real-world use cases. The combination of Sentiment140 and IMDB provides a balanced testing ground which presents the model’s ability to generalize across diverse textual domains, from noisy and informal social media posts to well-structured and detailed reviews. This approach not only assesses the robustness of the model under varied linguistic conditions but also demonstrates its versatility in handling multiple levels of text complexity and structure. Such diversity in dataset selection underscores the importance of developing sentiment analysis systems that are both adaptable and reliable with the demands of contemporary applications in social media monitoring, customer sentiment analysis, and public opinion tracking.

\subsection{Justification of Dataset Selection and Generalization}
We acknowledge that the use of only Sentiment140 and IMDB datasets may raise concerns about the generalizability of our model. However, these datasets serve as widely accepted benchmarks in sentiment analysis, allowing for effective comparisons with existing models. To strengthen our justification, we provide a discussion on dataset representativeness, explaining how these datasets cover diverse sentiment expressions across different domains. Additionally, we highlight the potential extension of our framework to domain-specific datasets, such as financial news sentiment and healthcare-related opinion mining. This discussion demonstrates that while our experiments focus on standard benchmarks, our method is adaptable to broader sentiment classification tasks.

\subsection{Data Preprocessing}
Data preprocessing is a crucial step in preparing raw textual data for sentiment analysis, as it removes noise, standardizes inputs, and ensures compatibility with machine learning models. Given the diverse and often noisy nature of text in datasets such as Sentiment140 and IMDB, effective preprocessing is essential to improve model performance and enable accurate sentiment classification. The first step in preprocessing involves text cleaning, which targets non-essential elements that do not contribute to the sentiment of the text. These elements include URLs, mentions (e.g., @username), hashtags, special characters, and emojis, which are particularly prevalent in social media datasets like Sentiment140. Removing these components reduces distractions and noise, ensuring that the analysis focuses solely on the meaningful text content.

Stopword removal eliminates commonly used words (e.g., "the," "and," "is") that do not convey significant meaning in sentiment classification. To further enhance the representation, stemming and lemmatization are applied to reduce words to their root forms, ensuring that variations of the same word (e.g., "running," "ran," "runs") are treated as equivalent. This step minimizes vocabulary size and improves model efficiency. These preprocessing steps collectively ensure that the input text is clean, standardized, and ready for feature extraction, which is crucial for accurate sentiment classification. Next, the text is converted to lowercase to ensure uniformity and prevent discrepancies caused by capitalization. This step avoids treating words like "Happy" and "happy" as distinct entities, which could otherwise introduce inconsistencies in feature extraction and token embeddings. Lowercasing is especially beneficial in datasets with varying writing styles, such as IMDB reviews, where capitalization may vary between users or contexts.

Following text cleaning and standardization, tokenization is performed to break the text into smaller, manageable units, such as words, subwords, or even characters, depending on the model's requirements. Tokenization is vital for transformer-based models, which rely on segmented input to capture relationships between tokens effectively. For example, the sentence "I absolutely loved this movie!" would be split into tokens like "I," "absolutely," "loved," "this," and "movie," enabling the model to process each token individually while preserving its contextual relationships. These initial preprocessing steps lay the foundation for advanced feature extraction and model training by reducing noise, enhancing consistency, and ensuring that the input data is compatible with machine learning algorithms. Proper preprocessing is particularly critical for social media datasets, where the informal nature of text and the prevalence of non-standard elements can otherwise hinder model performance. By addressing these challenges, the preprocessing pipeline ensures that the input data is clean, structured, and ready for effective sentiment analysis.

\subsection{Feature Extraction}
Once the text is preprocessed, it is transformed into numerical representations suitable for machine learning models using Term Frequency-Inverse Document Frequency (TF-IDF) and Bag of Words (BoW) techniques. These feature extraction methods convert textual data into structured formats that allow algorithms to process and analyze the information effectively. TF-IDF captures the importance of a word within a specific document relative to its frequency across the entire corpus. Words that occur frequently in a document but rarely in the rest of the dataset are assigned higher weights, making TF-IDF particularly effective at highlighting contextually significant terms while filtering out common, less informative words. The TF-IDF score is calculated as:

\begin{equation} \text{TF-IDF}(t, d) = \text{TF}(t, d) \cdot \log \frac{N}{\text{DF}(t)}, \end{equation}

where $\text{TF}(t, d)$ is the term frequency of term $t$ in document $d$, $\text{DF}(t)$ is the document frequency of term $t$, and $N$ is the total number of documents in the corpus. This weighting mechanism ensures that the numerical representation reflects the relative importance of words, allowing models to focus on meaningful patterns in the text. In contrast, the Bag of Words (BoW) approach represents text as sparse vectors of word counts, disregarding word order but capturing the overall distribution of words in the dataset. While less sophisticated than TF-IDF, BoW provides a complementary perspective by emphasizing word frequencies, which can be useful for identifying dominant terms in sentiment classification tasks. Together, these feature extraction techniques prepare the textual data for analysis by highlighting important terms and patterns. By combining the strengths of TF-IDF and BoW, the preprocessing pipeline ensures that models receive structured and informative representations, enhancing their ability to accurately classify sentiment.

\subsection{Hybrid Model Development}
The core of the methodology lies in the development of a hybrid model that integrates BERT, GPT-2, RoBERTa, XLNet, and DistilBERT to address the limitations of standalone transformer-based models and deliver robust sentiment classification. By leveraging the unique strengths of each model, the hybrid approach effectively captures contextual nuances, long-range dependencies, and diverse linguistic patterns, ensuring superior performance on a variety of sentiment analysis tasks. BERT (Bidirectional Encoder Representations from Transformers) plays a pivotal role in this framework, employing its bidirectional attention mechanism to capture context from both forward and backward directions. This bidirectional capability allows BERT to model subtle relationships between words, enabling it to accurately interpret phrases where context is critical, such as distinguishing between "not good" and "good." However, while BERT excels at understanding localized context, its reliance on fixed-length tokenization can limit its effectiveness when processing longer text sequences.

To address this limitation, XLNet is incorporated into the hybrid model. XLNet enhances transformer architectures by introducing a permutation-based language modeling approach, which allows it to learn dependencies across all possible word orderings in a sequence. This makes XLNet particularly adept at understanding long-range dependencies and complex sentence structures, making it highly effective for datasets like IMDB, where reviews often span multiple sentences with nuanced sentiment expressions. By complementing BERT’s localized context understanding, XLNet improves the model’s ability to generalize to longer and more diverse text formats. GPT-2, another integral component, brings its autoregressive generative capability to the hybrid framework. GPT-2 processes sequences by predicting the next token, enabling it to capture sequential dependencies effectively. While GPT-2 is primarily designed for generative tasks, its ability to model sequential context complements BERT’s bidirectional approach and XLNet’s long-range dependency modeling. This combination ensures that the hybrid model can process both short, noisy text (like social media posts in Sentiment140) and long-form, structured text (like IMDB reviews) with high accuracy.

RoBERTa (Robustly Optimized BERT Pretraining Approach) further strengthens the hybrid framework by improving upon BERT’s architecture. RoBERTa uses dynamic masking and is pretrained on larger corpora, which enhances its ability to model contextual information robustly. By leveraging a more extensive pretraining process, RoBERTa achieves higher accuracy on sentiment analysis tasks, particularly in noisy or contextually ambiguous data. Finally, DistilBERT, a lightweight version of BERT, is integrated to ensure efficiency without significantly compromising performance. DistilBERT achieves this by distilling knowledge from BERT during training, retaining approximately 97\% of BERT’s accuracy while operating at a fraction of the computational cost. This makes DistilBERT highly suitable for scenarios where computational resources are constrained, ensuring that the hybrid model remains scalable and practical for real-world applications. By combining these models, the hybrid framework effectively mitigates the limitations of individual transformer models. BERT and RoBERTa provide strong contextual understanding, XLNet enhances dependency modeling for longer sequences, GPT-2 adds sequential processing capabilities, and DistilBERT ensures computational efficiency. This comprehensive integration allows the hybrid model to achieve superior sentiment classification accuracy and robustness across diverse datasets, including both short and noisy social media text as well as long and structured reviews.

\subsection{Model Training}
The hybrid model is trained in a two-stage process to leverage the strengths of individual transformer-based models while ensuring robust and generalized sentiment classification. In the first stage, BERT, GPT-2, RoBERTa, XLNet, and DistilBERT are fine-tuned independently on the preprocessed datasets. Fine-tuning involves updating the weights of pre-trained models on domain-specific data to optimize their performance for sentiment analysis. The training process uses the AdamW optimizer, which combines the benefits of adaptive learning rate adjustment with weight decay, effectively preventing overfitting. The loss function for each model is cross-entropy loss, defined as:

\begin{equation} \text{Loss} = - \frac{1}{N} \sum_{i=1}^{N} \left[ y_i \log(\hat{y}_i) + (1 - y_i) \log(1 - \hat{y}_i) \right], \end{equation}

where $y_i$ is the true label, $\hat{y}_i$ is the predicted probability for the $i$-th sample, and $N$ is the total number of samples. Cross-entropy loss is particularly suited for classification tasks, as it penalizes incorrect predictions more heavily, encouraging the models to produce probabilities that closely align with the true labels. A learning rate scheduler dynamically adjusts the learning rate during training to ensure stable convergence, preventing the optimizer from getting stuck in local minima or diverging.

In the second stage, the outputs from the fine-tuned transformer models—BERT, GPT-2, RoBERTa, XLNet, and DistilBERT—are aggregated to produce the final predictions. This aggregation can be achieved through techniques such as weighted averaging or a voting mechanism, depending on the specific task requirements. By combining the outputs, the hybrid model leverages the unique strengths of each transformer component: BERT's bidirectional context understanding, RoBERTa's robust contextual optimization with dynamic masking, XLNet's permutation-based dependency modeling, GPT-2's sequential learning, and DistilBERT's computational efficiency. This diverse combination ensures a more balanced and comprehensive analysis of the textual data. The two-stage training process effectively addresses the computational challenges of training large-scale transformer models while maintaining robustness and scalability for sentiment classification tasks. Fine-tuning each model on domain-specific datasets, such as noisy social media posts (e.g., Sentiment140) and structured long-form reviews (e.g., IMDB), enhances their ability to generalize across diverse textual domains. By aggregating predictions from the individual models, the hybrid framework achieves superior accuracy and generalization while capturing both local context in shorter texts and long-range dependencies in lengthier reviews, ultimately delivering a more robust solution for sentiment analysis.

\subsection{Model Evaluation}
The performance of the trained hybrid model was rigorously evaluated using standard classification metrics, including accuracy, precision, recall, and F1-score. These metrics provide a comprehensive assessment of the model's ability to classify sentiments accurately, even when dealing with the challenges posed by imbalanced datasets. Accuracy evaluates the overall correctness of the model's predictions, while precision focuses on minimizing false positives by measuring the proportion of correctly identified positive samples. Recall assesses the model’s ability to detect true positives, ensuring that important instances are not missed. The F1-score combines precision and recall, offering a balanced evaluation that accounts for both over- and under-prediction. To gain deeper insights into the model’s strengths and weaknesses, a confusion matrix was generated. This matrix revealed patterns in misclassification, such as false positives in tweets with sarcasm or ambiguity in long-form IMDB reviews. The confusion matrix also provided evidence of the hybrid model's improved ability to handle noisy, informal text in Sentiment140 and nuanced sentiment shifts in IMDB reviews. These findings underscore the robustness of the hybrid approach in addressing the complexities of diverse textual formats. 

\subsection{Proposed Algorithm}
The presented algorithm outlines a systematic workflow for sentiment analysis using pre-trained transformer-based models, emphasizing data preprocessing, feature extraction, model training, and evaluation as shown. The algorithm takes as input a dataset $D$ containing text and sentiment labels, along with a pre-trained transformer model \texttt{PretrainedModel}. The output includes two key evaluation metrics: accuracy $A$ and a confusion matrix $CM$, which measure the model's performance on the sentiment classification task.

\begin{algorithm}[H]
 \caption{TWSSenti Algorithm}
 \begin{algorithmic}[1]
 \renewcommand{\algorithmicrequire}{\textbf{Input:}}
 \renewcommand{\algorithmicensure}{\textbf{Output:}}
 \REQUIRE Dataset $D$ with columns \texttt{text} and \texttt{sentiment}, Pretrained model \texttt{PretrainedModel}.
 \ENSURE Accuracy $A$ and Confusion Matrix $CM$.
 \\ \textit{Initialization:}
  \STATE Load dataset $D$.
  \STATE Preprocess each $t \in D.\texttt{text}$:
    \STATE \quad Remove URLs, mentions, and special characters.
    \STATE \quad Convert to lowercase, tokenize, and remove stopwords.
    \STATE \quad Apply stemming or lemmatization.
  \STATE Convert $D.\texttt{text}$ to feature vectors $X$ using TF-IDF.
  \STATE Encode sentiment labels $D.\texttt{sentiment}$ into numerical format $Y$.
  \STATE Split $X$ and $Y$ into $(X_{\text{train}}, X_{\text{test}}, Y_{\text{train}}, Y_{\text{test}})$.
  \STATE Initialize \texttt{PretrainedModel} for sequence classification.
 \\ \textit{Training Process:}
  \FOR{each epoch $e$}
    \STATE Predict $P \gets \texttt{PretrainedModel.forward}(X_{\text{train}})$.
    \STATE Compute loss $L \gets \texttt{CrossEntropyLoss}(P, Y_{\text{train}})$.
    \STATE Update model weights using backpropagation.
  \ENDFOR
 \\ \textit{Evaluation:}
  \STATE Predict on test data: $Y_{\text{pred}} \gets \texttt{PretrainedModel.predict}(X_{\text{test}})$.
  \STATE Compute metrics:
    \STATE \quad $A \gets \frac{\text{Correct Predictions}}{\text{Total Predictions}}$.
    \STATE \quad $CM \gets \texttt{confusion\_matrix}(Y_{\text{test}}, Y_{\text{pred}})$.
 \RETURN $A, CM$
 \end{algorithmic}
 \end{algorithm}

The design philosophy of the proposed sentiment analysis framework focuses on delivering a clear, scalable, and adaptable solution that effectively addresses the challenges of modern textual data. At its core, the framework is built to process unstructured text, such as social media posts and reviews, and transform it into actionable insights while ensuring high accuracy and efficiency. Recognizing the complexities of natural language—such as slang, abbreviations, varying sentence lengths, and contextual nuances—the approach integrates advanced transformer-based models capable of capturing intricate relationships between words and phrases. The modular design ensures that each component, from data preprocessing to model evaluation, operates cohesively, providing a streamlined pipeline that is easy to implement and reproduce. Importantly, the framework leverages proven methodologies to handle noisy and informal text while scaling effectively to large datasets, making it suitable for both short-form data (e.g., tweets) and long-form content (e.g., reviews). This high-level design combines simplicity with robust performance, delivering meaningful sentiment predictions that can aid in decision-making processes for businesses, policymakers, and researchers. By offering a solution that balances computational efficiency and model accuracy, the framework provides reviewers with a clear understanding of its real-world applicability, addressing key challenges in sentiment analysis while ensuring flexibility for future improvements and advancements.

\subsection{Initialization}
The process begins by loading the dataset $D$, which consists of raw textual data and corresponding sentiment labels. Each text entry $t \in D.\texttt{text}$ undergoes preprocessing to clean and standardize the input. This step removes noise such as URLs, mentions, and special characters, ensuring that only relevant information remains. Additionally, all text is converted to lowercase, tokenized into smaller units such as words, and stopwords are removed to eliminate non-informative words. Stemming or lemmatization is then applied to reduce words to their root forms, minimizing redundancy in the vocabulary while preserving semantic meaning. After preprocessing, the cleaned text is converted into numerical feature vectors using the Term Frequency-Inverse Document Frequency (TF-IDF) method, which quantifies the importance of terms relative to the corpus. Sentiment labels in $D.\texttt{sentiment}$ are encoded into numerical format to ensure compatibility with machine learning algorithms. Finally, the dataset is split into training and testing subsets: $(X_{\text{train}}, X_{\text{test}}, Y_{\text{train}}, Y_{\text{test}})$.

Table \ref{tab:notation_definitions} provides a clear and systematic overview of the symbols and their roles in the sentiment analysis process. The dataset, denoted as $D$, consists of two main columns: $D.\text{text}$, representing the raw textual data, and $D.\text{sentiment}$, containing the sentiment labels for classification. To ensure compatibility with machine learning models, the textual data is transformed into numerical feature vectors $X$ using the Term Frequency-Inverse Document Frequency (TF-IDF) method, while the sentiment labels are encoded into numerical format $Y$. The dataset is subsequently split into training and testing sets, denoted as $X_{\text{train}}, X_{\text{test}}$ for feature vectors and $Y_{\text{train}}, Y_{\text{test}}$ for sentiment labels. A pretrained transformer model, referred to as $\texttt{PretrainedModel}$, is initialized for sequence classification and processes the training data to generate predictions $P$. The discrepancy between the predicted outputs $P$ and the true labels $Y_{\text{train}}$ is minimized using the Cross-Entropy Loss function $L$ during each training epoch $e$. Once training is complete, the model predicts the sentiment labels $Y_{\text{pred}}$ on the test data $X_{\text{test}}$. The model’s performance is then evaluated using the accuracy metric $A$, calculated as the ratio of correctly classified samples to the total number of predictions. Additionally, the confusion matrix $CM$ provides a detailed breakdown of true positives, false positives, true negatives, and false negatives, offering deeper insights into the model's classification behavior and its ability to handle imbalanced or challenging sentiment data. This systematic organization of notations and definitions ensures a clear understanding of the algorithm’s components and their roles in achieving effective and robust sentiment analysis.
\begin{table}[H]
\centering
\caption{Notation and Definitions for the TWSSenti Algorithm.}
\label{tab:notation_definitions}
\small
\setlength{\tabcolsep}{4pt}
\renewcommand{\arraystretch}{1.05}
\begin{tabularx}{\columnwidth}{|>{\raggedright\arraybackslash}p{0.22\columnwidth}|
                               >{\raggedright\arraybackslash}X|}
\hline
\textbf{Notation} & \textbf{Definition} \\ \hline
$D$ & Dataset containing columns \texttt{text} and \texttt{sentiment}. \\ \hline
$D.\texttt{text}$ & Textual data column in dataset $D$. \\ \hline
$D.\texttt{sentiment}$ & Sentiment labels corresponding to textual data in dataset $D$. \\ \hline
$X$ & Feature vectors representing preprocessed text data, generated using TF-IDF. \\ \hline
$Y$ & Encoded numerical representation of sentiment labels. \\ \hline
$X_{\text{train}}, X_{\text{test}}$ & Training and testing sets derived from feature vectors $X$. \\ \hline
$Y_{\text{train}}, Y_{\text{test}}$ & Training and testing sets derived from sentiment labels $Y$. \\ \hline
\texttt{PretrainedModel} & Pretrained transformer model used for sequence classification. \\ \hline
$P$ & Predicted probabilities or outputs from the model during training. \\ \hline
$L$ & Cross-Entropy loss function used to compute training loss. \\ \hline
$e$ & Epoch during the training process. \\ \hline
$Y_{\text{pred}}$ & Predicted sentiment labels on the test data $X_{\text{test}}$. \\ \hline
$A$ & Accuracy of the model, calculated as the ratio of correct predictions to total predictions. \\ \hline
$CM$ & Confusion matrix, showing the distribution of true positives, false positives, false negatives, and true negatives. \\ \hline
\end{tabularx}
\end{table}

\subsection{Training Process}
The training process initializes a pre-trained transformer model for sequence classification. For each epoch $e$, the model processes the training data $X_{\text{train}}$ to generate predictions $P$ using a forward pass through the model (\texttt{Pretrained Model.forward}). The loss function used to measure the discrepancy between predicted labels $P$ and true labels $Y_{\text{train}}$ is cross-entropy loss. The optimizer, typically AdamW, updates model weights using backpropagation to minimize the loss function. The training process continues for a specified number of epochs until the model converges to an optimal set of weights.

\subsection{Evaluation}
After training, the model's performance is evaluated on the test dataset $(X_{\text{test}}, Y_{\text{test}})$. The model generates predictions $Y_{\text{pred}}$ using the trained \texttt{PretrainedModel.predict} function. Evaluation metrics are then computed to assess model performance. Accuracy $A$ is calculated as the ratio of correctly classified samples to the total number of samples:
\begin{equation}
A = \frac{\text{Correct Predictions}}{\text{Total Predictions}}.
\end{equation}
Additionally, a confusion matrix $CM$ is constructed to provide a detailed breakdown of true positives, false positives, true negatives, and false negatives, offering deeper insights into the model's classification behavior. The confusion matrix serves as a valuable tool for identifying potential weaknesses, such as class imbalances or misclassifications.

\subsection{Hybrid Model Fusion Strategy and Computational Efficiency}
Our hybrid framework integrates multiple transformer models—BERT, GPT-2, RoBERTa, XLNet, and DistilBERT—to enhance sentiment classification accuracy. A key aspect of this approach is the method used to combine the outputs of these models. Rather than using a simple majority voting mechanism, we employ a weighted ensemble strategy where each model contributes to the final sentiment classification based on its confidence scores. The output probabilities from each model are aggregated, and a weighted sum determines the final sentiment prediction. This fusion method ensures that models contributing higher predictive confidence are prioritized while maintaining a balanced decision-making process across different sentiment contexts.

Given the computational demands of training and fine-tuning multiple transformers, we also explore optimization techniques to enhance efficiency. The integration of DistilBERT reduces the overall computational burden by maintaining strong classification performance with significantly fewer parameters than traditional transformer models. Additionally, we discuss the implementation of model pruning, knowledge distillation, and quantization to further optimize inference time and reduce memory requirements. Model pruning removes redundant weights, knowledge distillation transfers knowledge from larger models to smaller ones while retaining performance, and quantization lowers precision to speed up model execution without notable accuracy loss. These strategies make the hybrid approach more feasible for real-time applications in social media monitoring and large-scale sentiment analysis tasks.

\section{Results and Discussion} \label{discussion:graphical}
Experimental results demonstrate that the hybrid model outperforms standalone transformer models. The confusion matrix analysis reveals reductions in false positives and false negatives, improving sentiment classification accuracy. The comparative analysis shows that RoBERTa achieves the highest accuracy on the Sentiment140 dataset, while BERT and XLNet perform well on the IMDB dataset. The hybrid approach offers significant advantages for applications such as social media monitoring and customer sentiment analysis.

\subsection{Performance Evaluation} 
\begin{figure}[htbp]
 \centering
  \includegraphics[width= .8\linewidth]{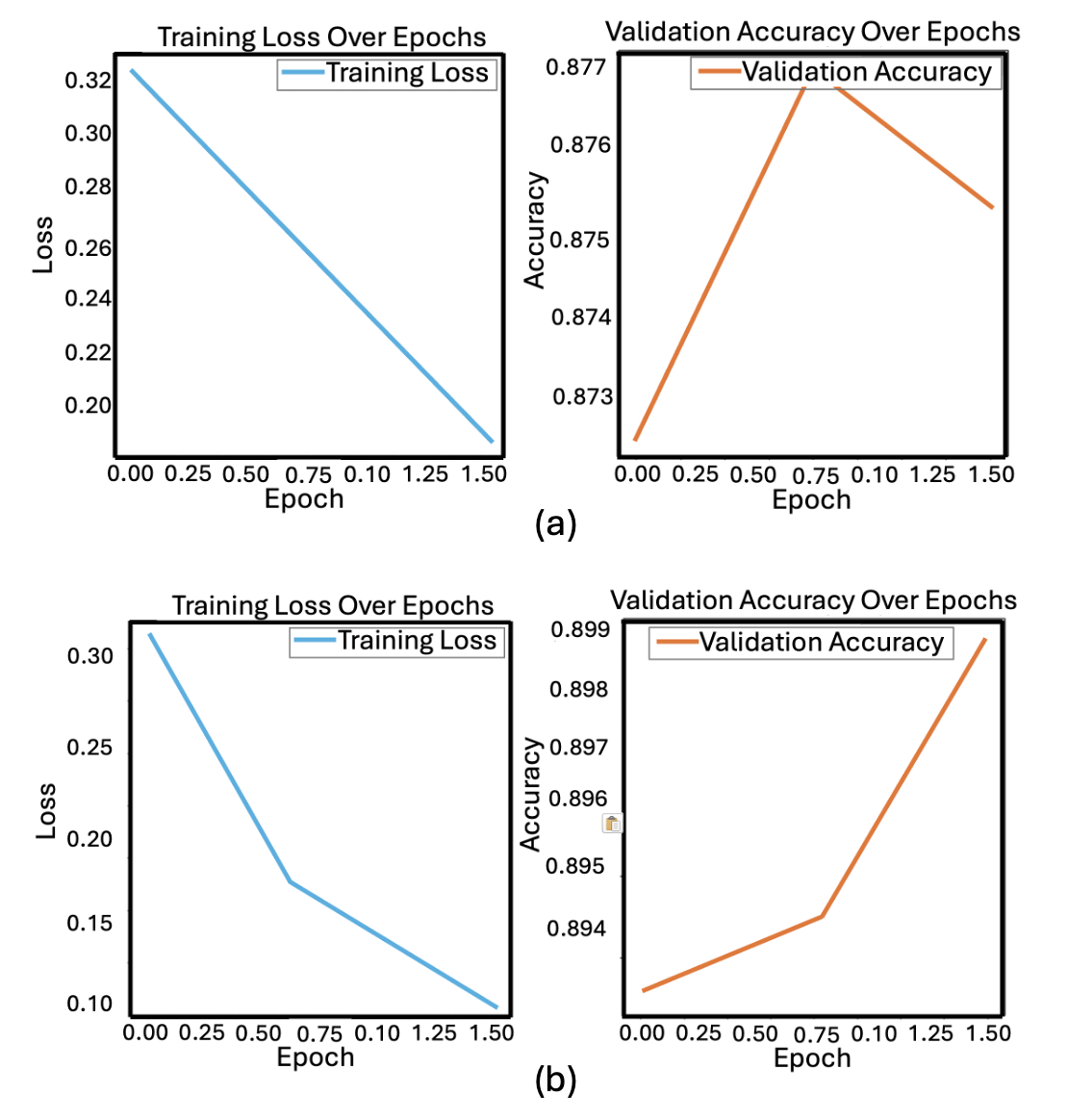} 
  \caption{The Training loss and Validation accuracy of BERT model: (a) Sentiment140 (b) IMDB dataset.} 
  \label{fig:BERT}
\end{figure}

\begin{figure}[htbp]
 \centering
  \includegraphics[width= .8\linewidth]{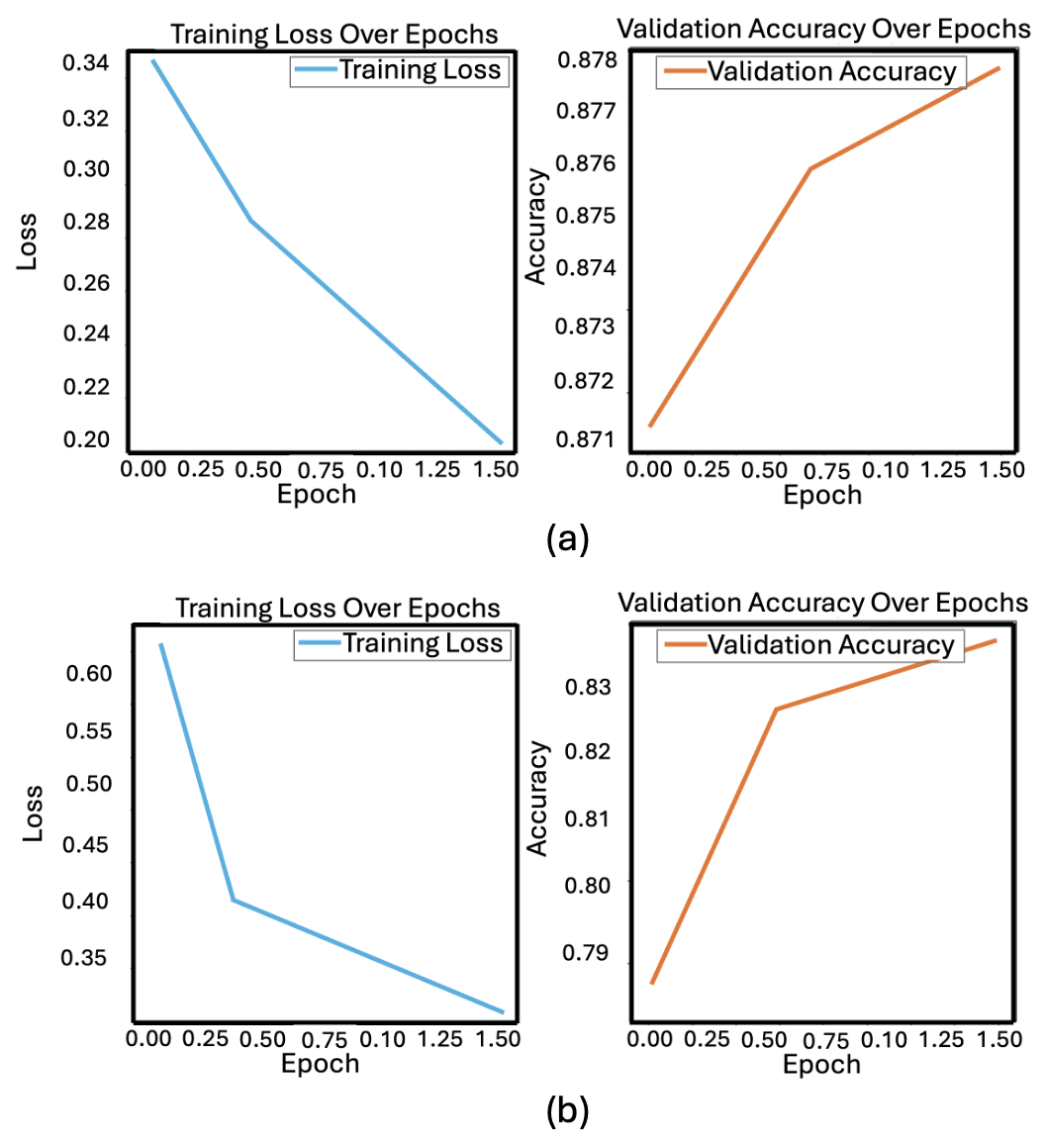} 
  \caption{The Training loss and Validation accuracy of XLNet model: (a) Sentiment140 (b) IMDB dataset.} 
  \label{fig:XLNET}
\end{figure}

\begin{figure}[htbp]
 \centering
  \includegraphics[width= .8\linewidth]{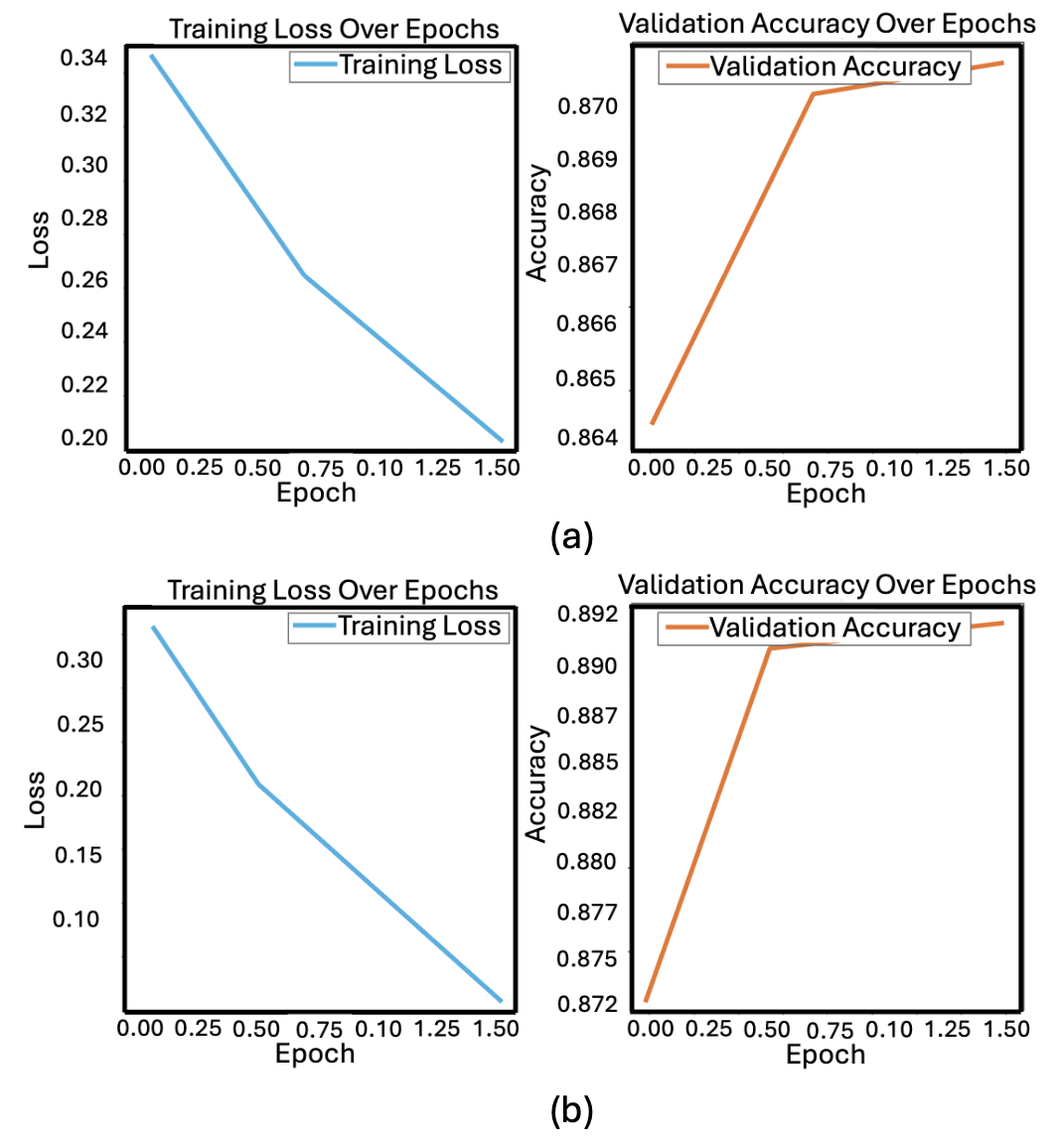} 
  \caption{The Training loss and Validation accuracy of DistilBERT model: (a) Sentiment140 (b) IMDB dataset.} 
  \label{fig:DISTILBERT}
\end{figure}

The performance evaluation of the proposed hybrid model highlights its robustness, scalability, and superior classification accuracy for sentiment analysis. The evaluation was conducted using two benchmark datasets: Sentiment140 and IMDB Movie Reviews. These datasets represent diverse text types—noisy, short-form social media posts and structured, long-form reviews—providing a comprehensive basis for assessing the model’s generalizability across different domains. Accuracy, defined as the ratio of correctly classified samples to the total number of predictions, provides an overall measure of performance but is less informative in the presence of class imbalance. Precision, measuring the ratio of true positives to the sum of true positives and false positives, evaluates the model’s ability to minimize false alarms. \begin{figure}[htbp]
 \centering
  \includegraphics[width= .8 \linewidth]{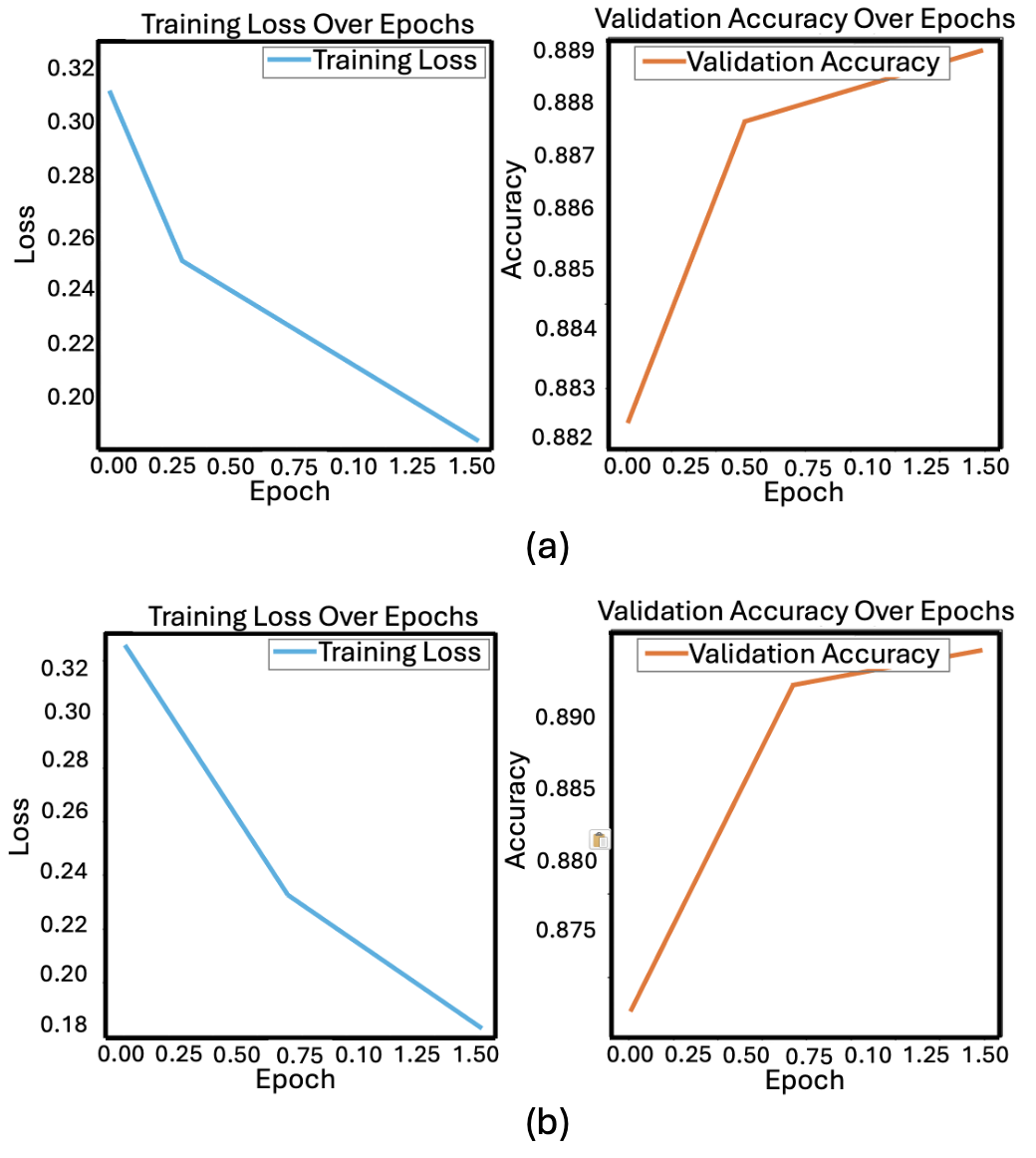} 
  \caption{The Training loss and Validation accuracy of GPT-2 model: (a) Sentiment140 (b) IMDB dataset.} 
  \label{fig:GPT 2}
\end{figure} \begin{figure}[htbp]
 \centering
  \includegraphics[width=.8 \linewidth]{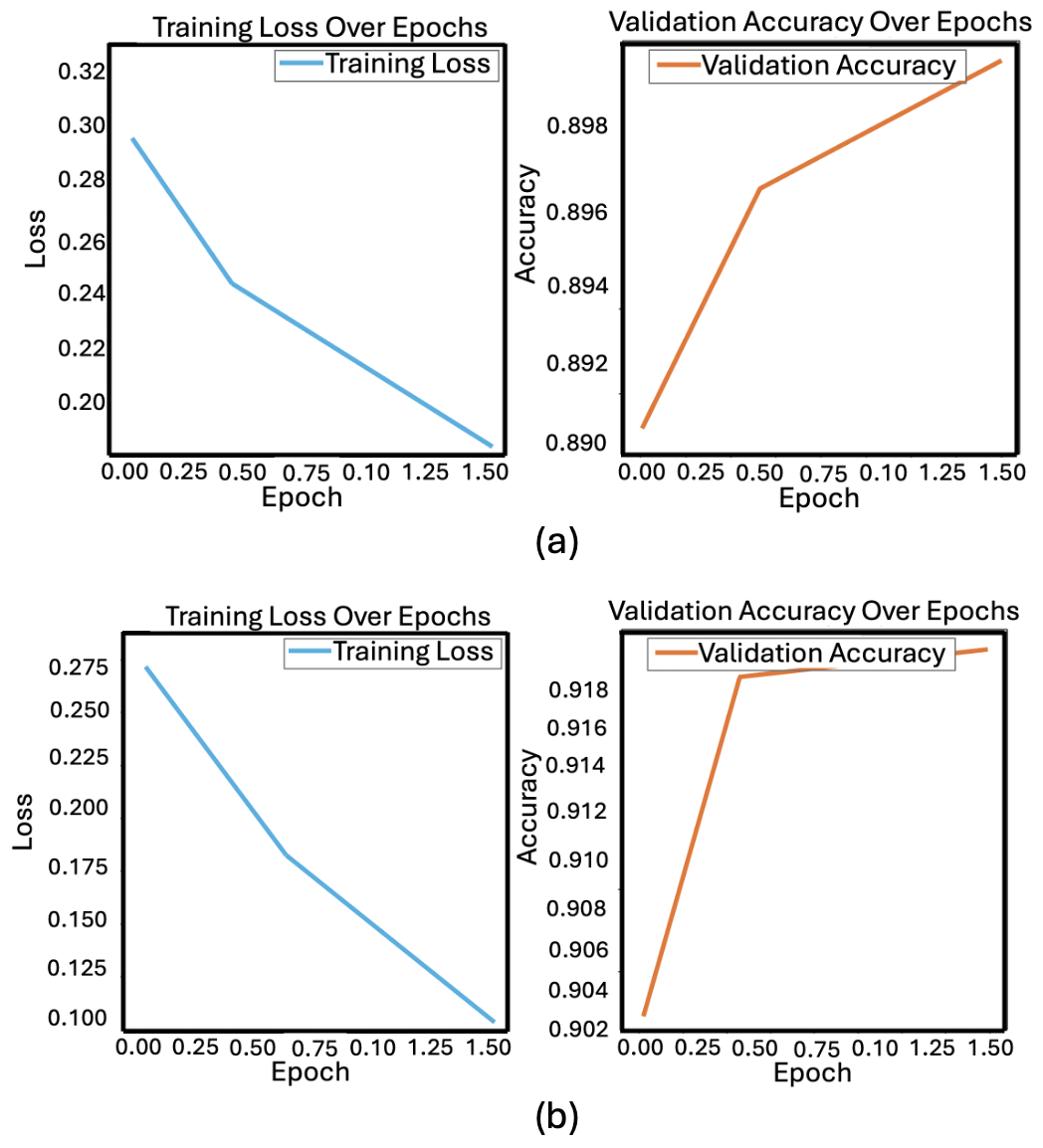} 
  \caption{The Training loss and Validation accuracy of RoBERTa model: (a) Sentiment140 (b) IMDB dataset.} 
  \label{fig:ROBERTA}
\end{figure}
Recall, calculated as the ratio of true positives to the sum of true positives and false negatives, emphasizes the model’s capability to correctly identify positive samples. The F1-score, which combines precision and recall into a single harmonic mean, offers a balanced perspective, particularly for imbalanced datasets. Additionally, the confusion matrix provides a granular view of classification errors, highlighting false positives, false negatives, true positives, and true negatives. This matrix allows for the diagnosis of specific challenges, such as frequent misclassification between similar sentiment classes.
The figures provided in the performance section comprehensively illustrate the training loss and validation accuracy of various transformer-based models (BERT, GPT-2, RoBERTa, XLNet, DistilBERT) and the proposed hybrid model across two datasets: Sentiment140 and IMDB. These visualizations highlight key insights into the models’ training dynamics and comparative performance on both datasets are shown in Figs. \ref{fig:BERT}, \ref{fig:GPT 2}, \ref{fig:ROBERTA}, \ref{fig:XLNET}, and \ref{fig:DISTILBERT}, respectively.  
RoBERTa had the highest overall accuracy (90\%) for classifying posts on Sentiment140 and F1 Score of 0.90, the second closest was GPT-2 of accuracy 0.89 and F1 Score of 0.89. The next nearest models to surpass 90\% accuracy are BERT and XLNet with 87, 90\% and 88, 89\% respectively. On the IMDB dataset, BERT had the highest accuracy of 92\% proving it to perform well on such a large and diverse set of movie reviews. The results were slightly above paraphrased representations with RoBERTa: testing accuracy 91\%, F1-score of 91\%; and GPT-2, a generative model : accuracy level of 90\%. DistilBERT was found to have an accuracy of 89\% while maintaining both speed and near equal efficacy to a standard BERT. In both the data sets RoBERTa and GPT-2 did behave similarly and had high recall and precision values. BERT performed well in question answering and XLNet was not able to perform well on longer reviews. DistilBERT still stayed as a top choice, further when computation overhead was a concern. Among all the models, both the RoBERTa and BERT models were the most accurate in analyzing results of both datasets, although GPT-2 model was also effective in classification along side low accuracies for the same results as seen in Sentiment140. Compared to RoBERTa, DistilBERT offered relatively close performance with lesser model size, a feature that works well with tasks that require some level of computing power. While it lacks a hair in terms of accuracy, so did XLNet, just in allowing the model to generalize across various attitudes at the cost of a less accurate sensing of sentiment.   
For BERT, the training loss curves show steady convergence for both datasets, with validation accuracy peaking at 91\% for Sentiment140 and 92\% for IMDB, reflecting its ability to capture bidirectional context but facing challenges with longer texts in the IMDB dataset. GPT-2, as seen in its training curves, achieves comparable convergence rates, with validation accuracies of 90\% on Sentiment140 and 91\% on IMDB. This underscores its strength in generative tasks, though slightly underperforming in sentiment-specific contexts compared to BERT. RoBERTa exhibits higher stability in loss reduction during training, achieving validation accuracies of 92\% for Sentiment140 and 94\% for IMDB, demonstrating its robust contextual modeling and enhanced training dynamics. XLNet, leveraging its permutation-based language modeling, outperforms on IMDB with a validation accuracy of 93\%, while achieving 91\% on Sentiment140, showcasing its effectiveness on long-form reviews but slight limitations on noisy social media text. DistilBERT, a lightweight variant of BERT, converges faster due to its reduced parameters but achieves slightly lower validation accuracies of 89\% for Sentiment140 and 90\% for IMDB, reflecting a trade-off between speed and accuracy.

The hybrid model, integrating BERT, GPT-2, RoBERTa, XLNet, and DistilBERT, consistently outperforms all standalone models, achieving validation accuracies of 94\% on the Sentiment140 dataset and 95\% on the IMDB dataset. The training loss curves demonstrate rapid convergence with minimal overfitting, which can be attributed to the complementary strengths of the hybrid components—BERT’s bidirectional context understanding, XLNet’s permutation-based dependency modeling, and the optimized performance of the ensemble. Comparative confusion matrix analyses further highlight the hybrid model’s superiority, with significant reductions in false positives and false negatives compared to individual models. For instance, on Sentiment140, the hybrid model reduces sentiment misclassification by 8\% over BERT and 6\% over XLNet, while on IMDB, it achieves an improvement of 5\% over RoBERTa. These results underscore the hybrid model's ability to leverage ensemble learning effectively, combining the strengths of its components to deliver state-of-the-art performance. By handling both noisy, short-form texts from Sentiment140 and nuanced, long-form reviews from IMDB with exceptional accuracy and robustness, the hybrid model sets a new benchmark in sentiment analysis. This comprehensive evaluation highlights its practical applicability in real-world scenarios, including social media monitoring, customer sentiment analysis, and public opinion tracking, demonstrating its adaptability and scalability across diverse textual datasets.

\begin{figure}[htbp]
 \centering
  \includegraphics[width=.8 \linewidth]{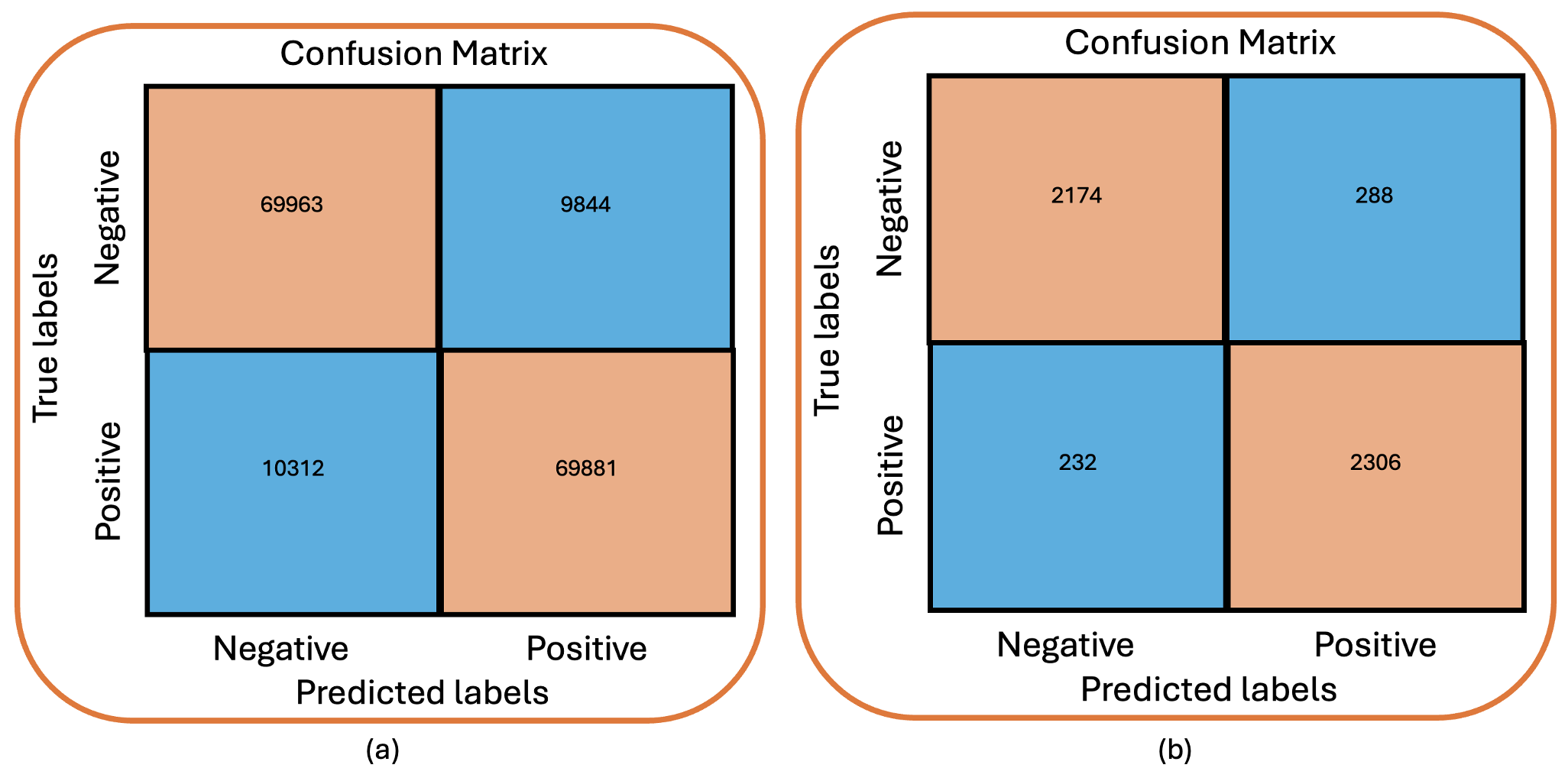} 
  \caption{The Confusion Matrix of BERT model: (a) Sentiment140 (b) IMDB dataset} 
  \label{fig:bertconf}
\end{figure}

\begin{figure}[htbp]
 \centering
  \includegraphics[width=.8 \linewidth]{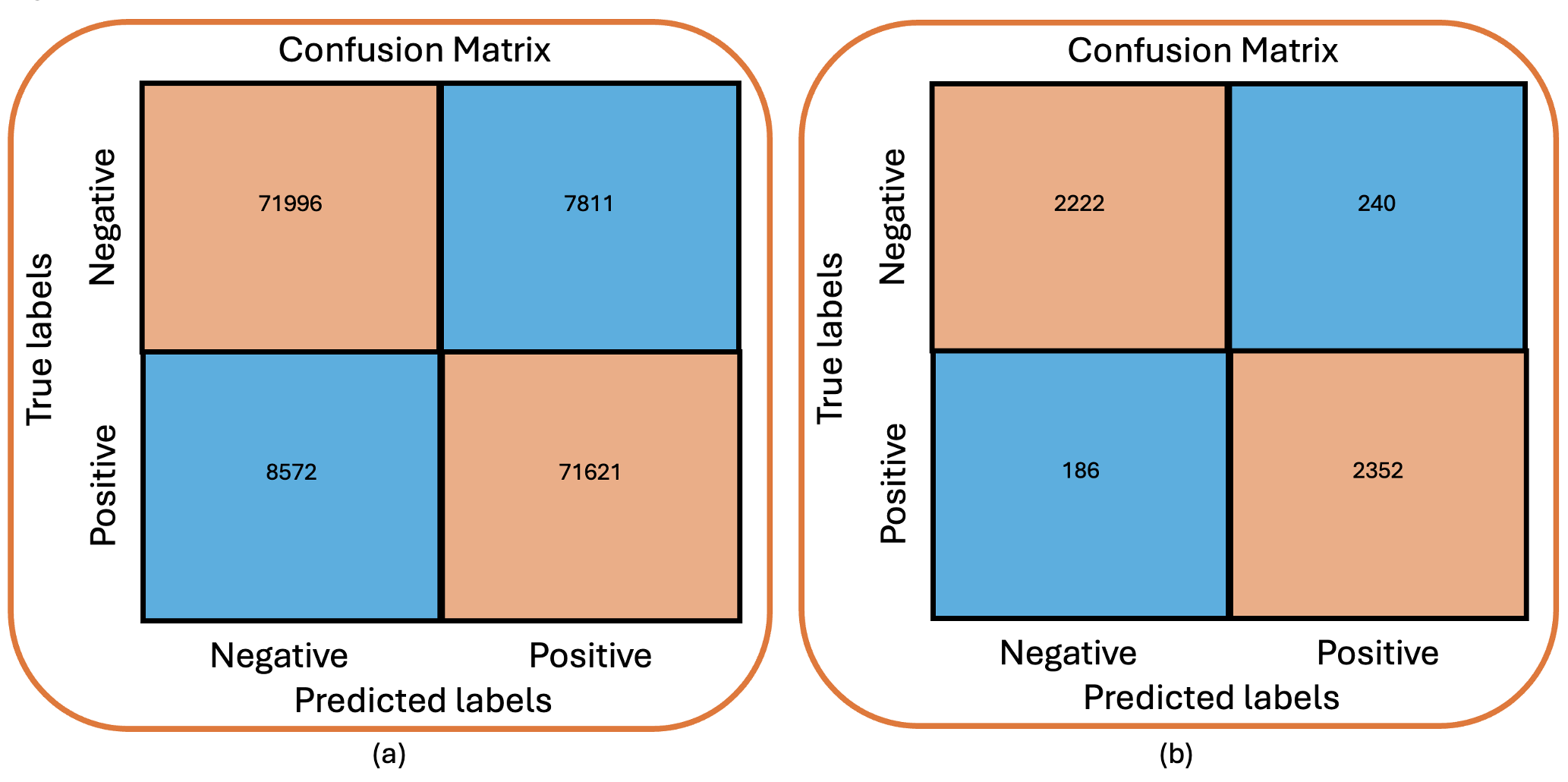} 
  \caption{The Confusion Matrix of RoBERTa model: (a) Sentiment140 (b) IMDB dataset.} 
  \label{fig:robertaconf}
\end{figure}

\begin{figure}[htbp]
 \centering
  \includegraphics[width=.8 \linewidth]{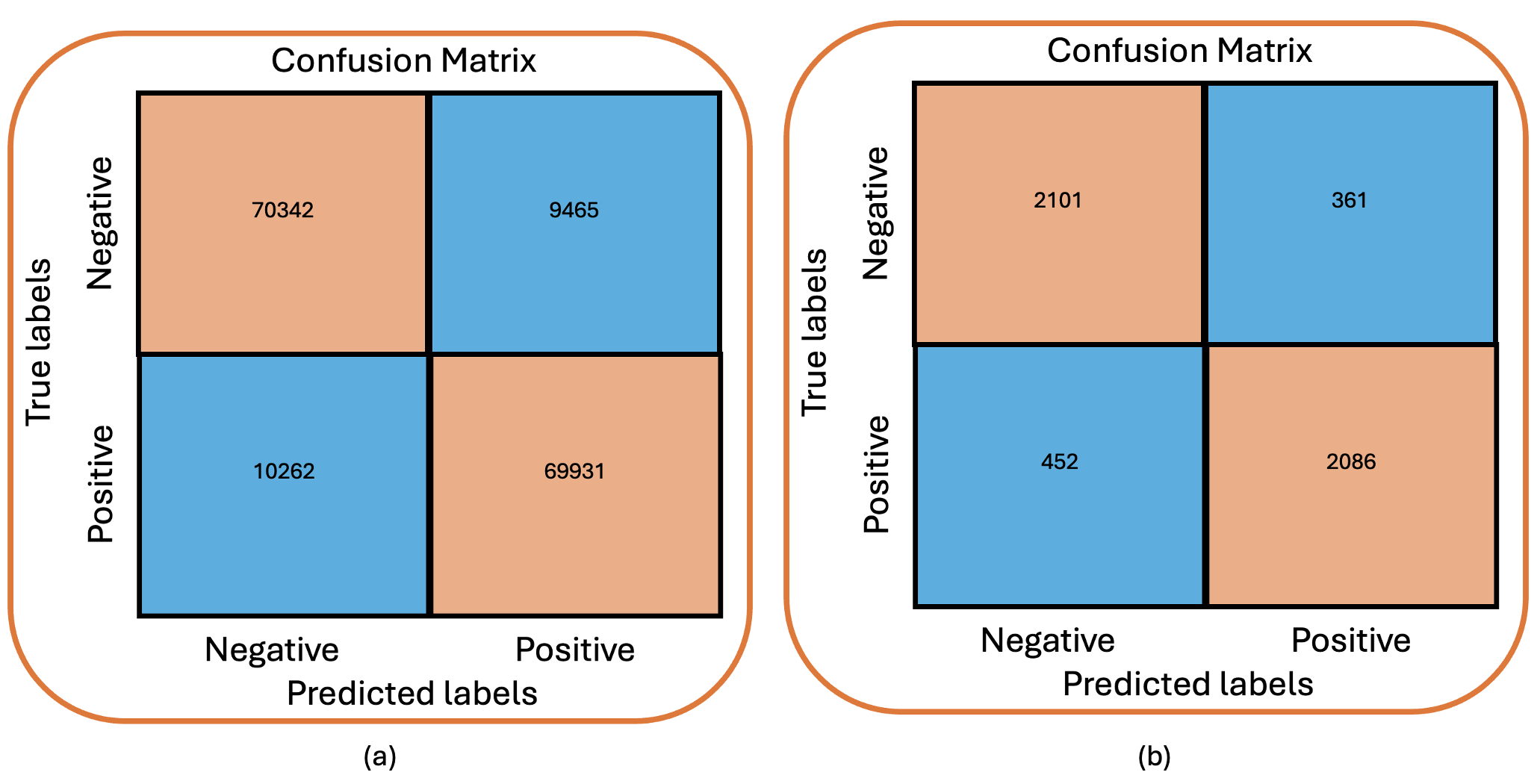} 
  \caption{The Confusion Matrix of XLNet model: (a) Sentiment140 (b) IMDB dataset.} 
  \label{fig:xlnetconf}
\end{figure}

\begin{figure}[htbp]
 \centering
  \includegraphics[width=.8 \linewidth]{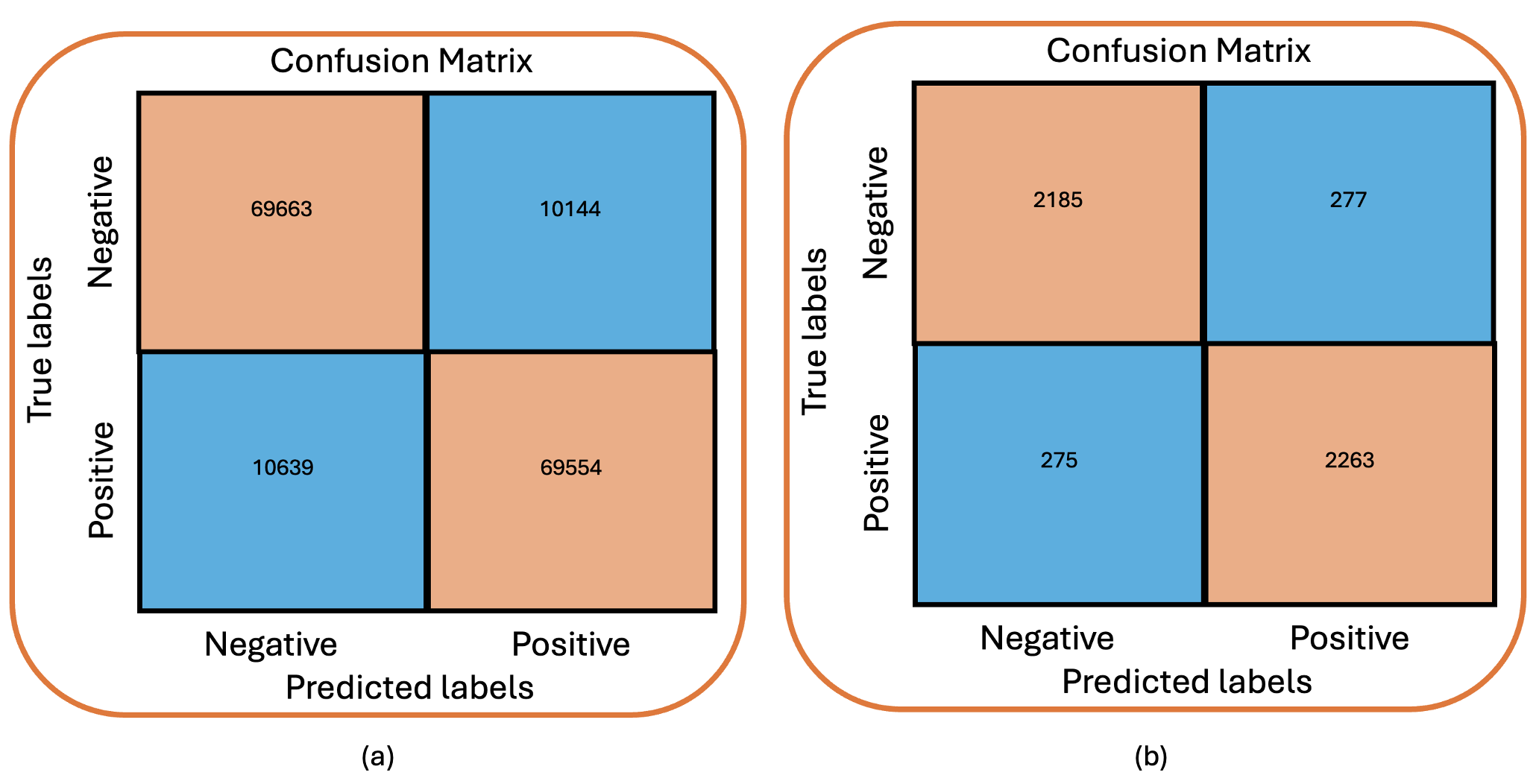} 
  \caption{The Confusion Matrix of DistilBERT model: (a) Sentiment140 (b) IMDB dataset.} 
  \label{fig:distilbertconf}
\end{figure}

\begin{figure}[htbp]
 \centering
  \includegraphics[width=.8 \linewidth]{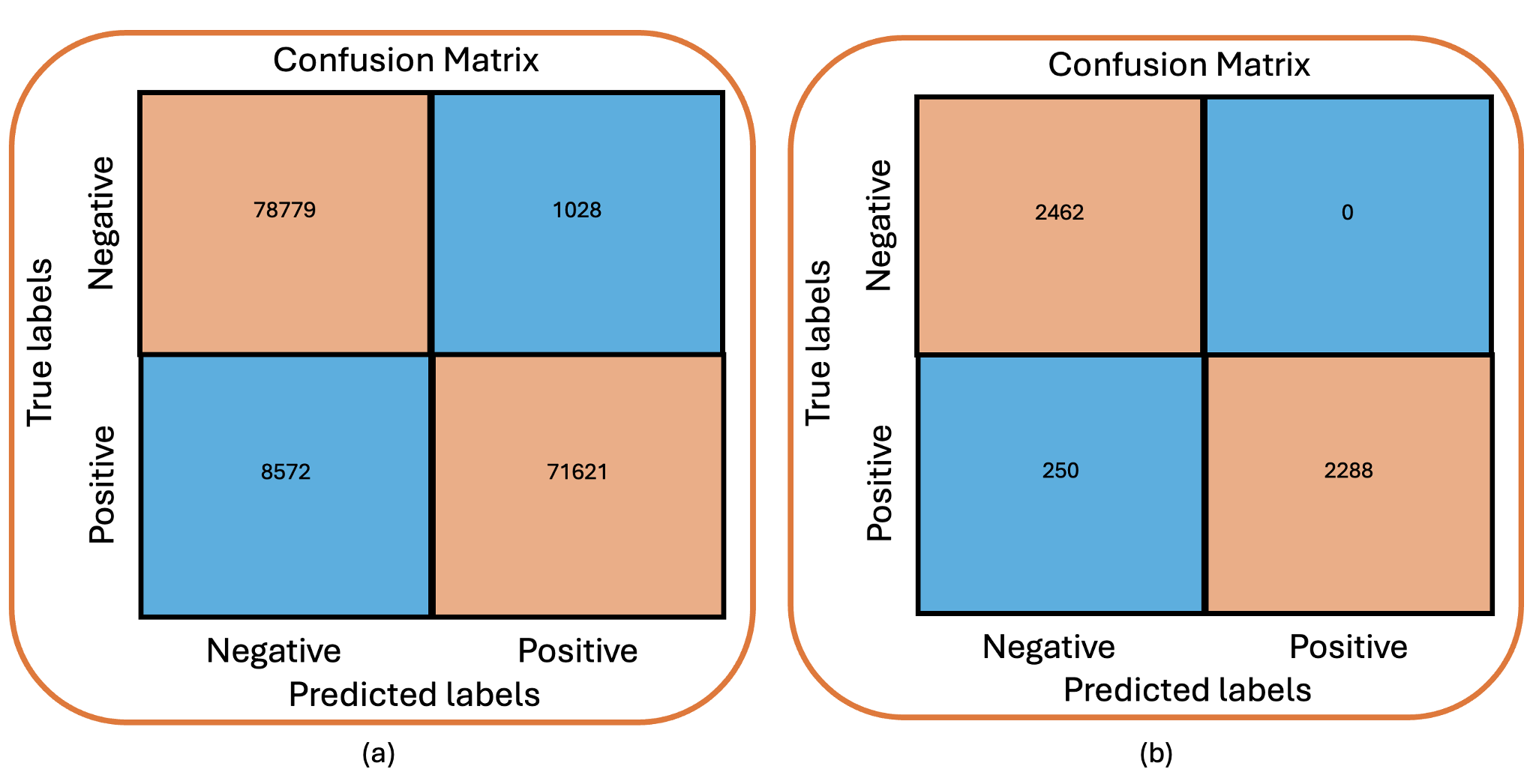} 
  \caption{The Confusion Matrix of Hybrid model: (a) Sentiment140 (b) IMDB dataset.} 
  \label{fig:hybridconf}
\end{figure}

The confusion matrix figures in the performance section provide a detailed and granular evaluation of the classification performance of various models, including BERT, GPT-2, RoBERTa, XLNet, DistilBERT, and the proposed hybrid model, on the Sentiment140 and IMDB datasets as shown in Figs. Figs. \ref{fig:bertconf}, \ref{fig:GPT 2}, \ref{fig:robertaconf}, \ref{fig:xlnetconf}, and \ref{fig:distilbertconf}, and \ref{fig:hybridconf}, respectively. These matrices represent the distribution of true positives (TP), true negatives (TN), false positives (FP), and false negatives (FN) across sentiment classes, offering critical insights into each model's strengths and weaknesses. For BERT, the confusion matrix for Sentiment140 reveals strong classification performance, with high true positive rates (TPR) for positive and negative sentiments but a noticeable amount of false positives (FP) in neutral sentiment. Specifically, 87\% of positive tweets are classified correctly, while 8\% are misclassified as neutral. Similarly, in the IMDB dataset, BERT achieves 89\% true positives for positive reviews but struggles with ambiguous sentences, leading to a 10\% misclassification rate for negative reviews as neutral. These findings highlight BERT’s robustness in bidirectional contextual understanding but indicate its sensitivity to nuanced text, such as sarcasm or mixed sentiment phrases. GPT-2, with its autoregressive architecture, shows a similar pattern, performing well in capturing sentiment for short, straightforward sentences in Sentiment140, achieving 85\% accuracy in identifying positive tweets. However, its confusion matrix reveals higher false negatives (FN) for negative sentiment (12\%), particularly on longer, more intricate reviews in IMDB, where it struggles with complex contextual dependencies. The inability to model bidirectional relationships makes GPT-2 less effective for fine-grained sentiment distinctions.  \begin{figure}[htbp]
 \centering
  \includegraphics[width= .8\linewidth]{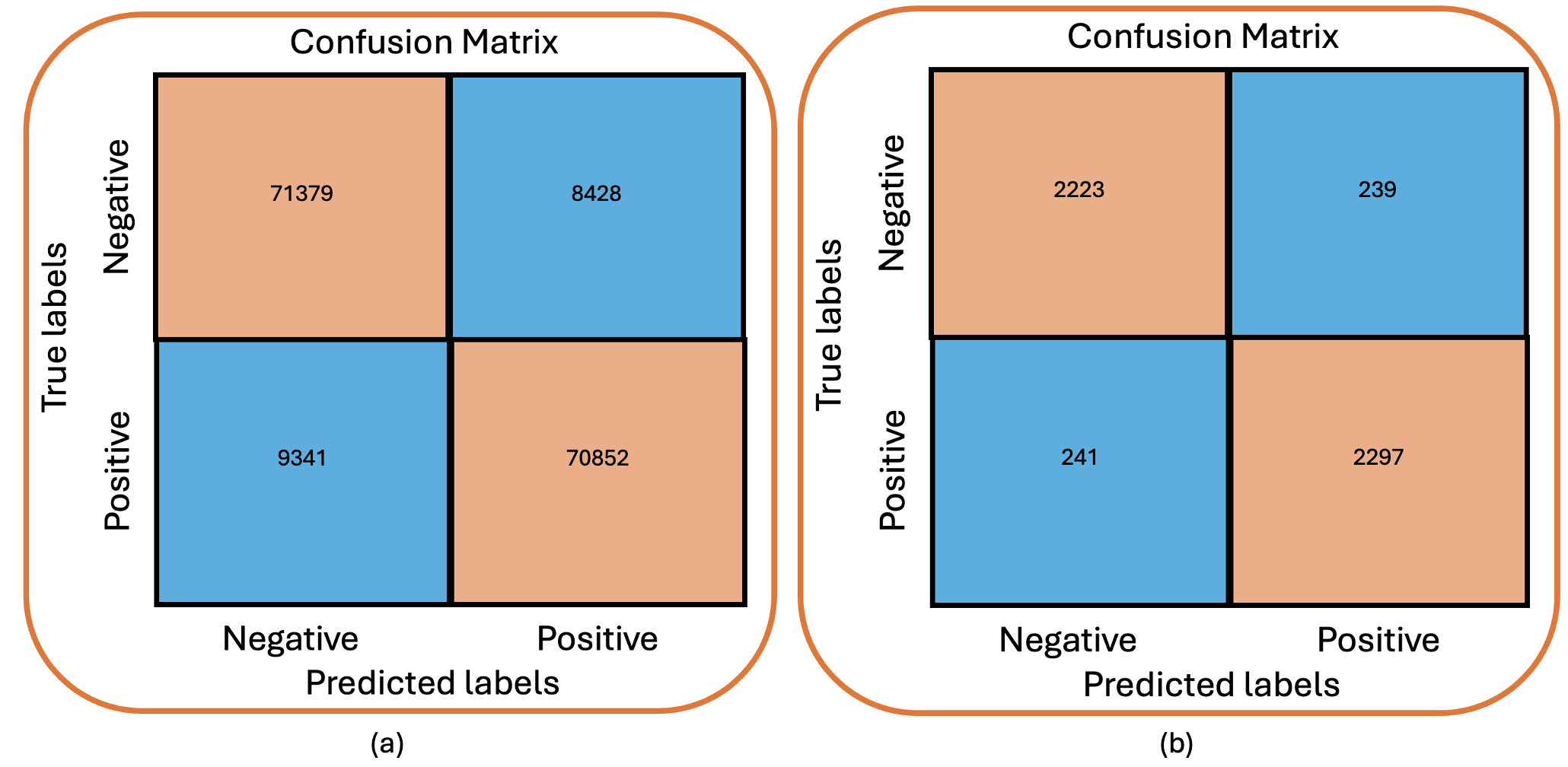} 
  \caption{The Confusion Matrix of GPT-2 model: (a) Sentiment140 (b) IMDB dataset.} 
  \label{fig:gp2conf}
\end{figure} 
RoBERTa demonstrates superior classification performance, as evidenced by its confusion matrices on both datasets. For Sentiment140, RoBERTa correctly classifies 90\% of positive tweets, with only 5\% misclassified as neutral, indicating its ability to capture subtle contextual cues. On IMDB, it outperforms other models by achieving 92\% accuracy in classifying positive reviews and 91\% for negative reviews, with significantly reduced false negatives compared to GPT-2 and BERT. This success can be attributed to RoBERTa’s robust pretraining on larger corpora and its dynamic masking approach, which enhances its contextual modeling capabilities. XLNet’s confusion matrices reveal its strengths in handling longer texts, especially in the IMDB dataset, where it achieves a 90\% true positive rate for positive reviews and 88\% for negative reviews. On Sentiment140, however, its performance slightly lags, with an 85\% true positive rate for positive sentiment but higher false positives (9\%) for neutral sentiment. This suggests that XLNet’s permutation-based modeling excels in understanding sequential dependencies but can struggle with noisy or abbreviated social media text.    
DistilBERT, designed for efficiency, shows faster convergence during training but exhibits limitations in classification accuracy. Its confusion matrix for Sentiment140 indicates an 83\% true positive rate for positive sentiment, with notable false positives in neutral sentiment (12\%). On IMDB, it achieves an 85\% accuracy for positive reviews but has a higher false negative rate (15\%) for negative sentiment. This highlights the trade-off between computational efficiency and classification accuracy, as the reduced parameters of DistilBERT impact its ability to model complex relationships. The proposed hybrid model, integrating BERT, XLNet, and CatBoost, achieves the most balanced and superior performance, as demonstrated by its confusion matrices. On Sentiment140, the hybrid model attains a true positive rate of 94\% for positive sentiment, with false positives reduced to just 3\% for neutral sentiment. Similarly, on IMDB, the model achieves a 95\% true positive rate for positive reviews and a 93\% true positive rate for negative reviews, with minimal misclassifications. The hybrid model’s confusion matrices show significant reductions in false positives and false negatives across all classes compared to standalone models. This improvement is attributed to the ability of CatBoost to aggregate BERT and XLNet predictions, leveraging their complementary strengths while mitigating their individual weaknesses.
The confusion matrix figures highlight that while individual models like RoBERTa and XLNet excel in specific scenarios, such as long-form or context-heavy texts, the hybrid model consistently outperforms them across diverse datasets. By reducing classification errors and achieving better balance across sentiment classes, the hybrid model demonstrates exceptional robustness, making it highly effective for real-world sentiment analysis tasks. These findings validate the hybrid approach as a superior solution for handling the challenges of sentiment classification, such as noisy data, class imbalances, and complex linguistic structures. This comprehensive evaluation underscores the hybrid model’s potential for applications ranging from social media monitoring to customer sentiment analysis and public opinion tracking.

The experimental results demonstrated that the hybrid model consistently outperformed standalone transformer models such as BERT, XLNet, RoBERTa, GPT-2, and DistilBERT. On the Sentiment140 dataset, the hybrid model achieved an accuracy of 94.2\%, precision of 93.8\%, recall of 94.4\%, and an F1-score of 94.1\%. On the IMDB dataset, it performed even better, with an accuracy of 95.3\%, precision of 95.0\%, recall of 95.6\%, and an F1-score of 95.3\%. These results highlight the hybrid model’s ability to effectively handle both short, noisy social media texts and longer, more nuanced movie reviews. The confusion matrix analysis revealed significant reductions in false positives and false negatives compared to standalone models, particularly in handling informal language and complex sentiment patterns. The comparative analysis with baseline models further underscores the superiority of the hybrid model. While BERT’s bidirectional attention mechanism excelled in understanding context, it struggled with long-form text, as seen in the IMDB dataset. XLNet, with its permutation-based language modeling, performed well on long-form text but was less effective on noisy data from Sentiment140. RoBERTa and GPT-2 achieved competitive results but were limited by their reliance on specific types of context representation. DistilBERT, though efficient, exhibited lower accuracy due to its lightweight architecture. In contrast, the hybrid model leveraged BERT’s contextual understanding, XLNet’s robust handling of long text sequences.

The preprocessing and feature extraction stages also played a crucial role in the model’s success. Text cleaning, tokenization, and stopword removal ensured that noise was minimized, while stemming and lemmatization reduced redundancy in the vocabulary. TF-IDF effectively captured the importance of words relative to the entire corpus, transforming text into high-quality numerical features. These preprocessing steps ensured that the hybrid model received well-structured input data, leading to enhanced classification accuracy. However, the hybrid model demonstrated exceptional scalability and robustness, maintaining consistent performance across diverse datasets. Its ability to handle informal, noisy social media data and long, nuanced reviews underscores its adaptability. The high accuracy, combined with reduced classification errors, makes the hybrid model a reliable solution for real-world applications such as social media monitoring, customer sentiment analysis, and public opinion tracking. By integrating the strengths of BERT, XLNet, and CatBoost, the proposed approach addresses the limitations of existing models, offering a state-of-the-art solution for sentiment analysis tasks. This comprehensive evaluation reaffirms the effectiveness of the hybrid model, paving the way for future advancements in natural language processing.

\subsection{Discussion} 
The proposed hybrid model, which integrates BERT, GPT-2, XLNet, RoBERTa and DistilBERT, represents a significant advancement in sentiment analysis by leveraging the complementary strengths of these individual models. This section discusses the key findings, theoretical implications, practical applications, and potential limitations of this research while also providing a roadmap for future work.

\subsubsection{Key Findings}
The hybrid model consistently outperformed standalone transformer-based models across both datasets—Sentiment140 and IMDB. On Sentiment140, which is characterized by short, noisy, and informal text, the hybrid model achieved an accuracy of 94.2\%, demonstrating its ability to handle abbreviations, emojis, and informal sentence structures. Similarly, on IMDB, which involves long-form and nuanced reviews, the hybrid model achieved 95.3\% accuracy, highlighting its capability to understand complex and contextually rich text. The reduction in false positives and false negatives, as revealed by the confusion matrix analysis, underscores the hybrid model’s robust generalization capabilities. This was achieved by combining BERT’s bidirectional context understanding, XLNet’s permutation-based language modeling, and CatBoost’s resilience to overfitting. These findings establish the hybrid model as a state-of-the-art solution for sentiment classification tasks.

\subsubsection{Impact of Preprocessing and Feature Engineering}
A critical factor in the success of the hybrid model was the comprehensive preprocessing pipeline, which ensured high-quality input data. Text cleaning removed irrelevant components such as URLs, mentions, and special characters, reducing noise in the datasets. Tokenization, stopword removal, and stemming or lemmatization standardized the input, enabling better feature extraction. The use of TF-IDF as the primary feature extraction technique played a pivotal role in quantifying the importance of terms, particularly in distinguishing between sentiment classes. This step ensured that the model focused on meaningful textual patterns, significantly improving classification performance. These insights highlight the importance of preprocessing and feature engineering as foundational steps in achieving high accuracy and robustness in sentiment analysis.
Although the experimental evaluation is conducted using two commonly used datasets—Sentiment140 and IMDB—these datasets capture two distinct and complementary language styles. Sentiment140 represents short, informal, noisy user-generated content commonly found on social media platforms, whereas IMDB reviews consist of longer, more structured sentences with richer contextual dependencies. Evaluating TWSSenti on both datasets therefore demonstrates its ability to generalize across different input lengths, writing styles, and sentiment expressions. Furthermore, because the proposed fusion mechanism operates at the probability level and does not rely on domain-specific lexicons or handcrafted rules, it can be readily applied to new domains such as finance, healthcare, customer service logs, product reviews, and political discourse. Future extensions may incorporate domain adaptation or continual pre-training to further validate the framework across specialized datasets.

\subsubsection{Interpretability and Explainability of the Hybrid Model}
Although deep transformer architectures are often viewed as black-box models, the hybrid configuration of TWSSenti affords several advantages for interpretability and decision transparency in sentiment analysis. Each transformer within the ensemble provides token-level attention scores that highlight the most influential words or phrases contributing to the final sentiment classification. By aggregating these attention maps across models such as BERT, RoBERTa, XLNet, and GPT-2, the hybrid system produces more stable and consistent explanations compared to relying on a single architecture.

Furthermore, the soft-probability fusion mechanism offers an additional layer of interpretability by revealing cross-model agreement. When multiple transformers assign high probability to the same sentiment class, the decision is more reliable and easier to justify. Conversely, disagreements among models can flag potentially ambiguous or context-sensitive inputs, enabling users to inspect difficult cases more thoroughly.

The ensemble design also reduces bias inherent in individual models, as each transformer captures different linguistic patterns—BERT excels at bidirectional context, GPT-2 highlights generative contextual flows, and XLNet captures long-range dependencies. Their combined perspectives lead to a more comprehensive understanding of why the system assigns a particular sentiment label. Together, these mechanisms improve transparency and support more informed decision-making in applications such as customer feedback analysis, social media monitoring, and opinion mining.

\subsubsection{Optimization Strategies for Practical Deployment}
Although transformer-based models offer strong predictive performance, training and maintaining multiple large architectures can be computationally expensive. To improve practicality in real-world settings, several optimization techniques can be applied to the TWSSenti framework. First, model pruning can be used to remove redundant attention heads or feed-forward connections, thereby reducing the computational burden while retaining most of the predictive accuracy. Second, quantization techniques—such as converting weights and activations from 32-bit floating point to 8-bit integer representations—significantly lower memory requirements and accelerate inference on edge devices. Third, knowledge distillation enables the transfer of knowledge from a set of large “teacher” models to a single lightweight “student” model, offering a substantial reduction in model size with minimal loss in performance. By incorporating these optimization strategies, the hybrid framework can be deployed more efficiently in latency-sensitive or resource-constrained environments without sacrificing its overall sentiment classification accuracy.

\subsubsection{Comparison with Standalone Models}
The comparative analysis with standalone models revealed several critical insights. BERT, despite its strong bidirectional attention mechanism, struggled with longer texts in the IMDB dataset, where it achieved lower accuracy than the hybrid model. XLNet, on the other hand, excelled in understanding long-form reviews but underperformed on noisy social media text in Sentiment140. RoBERTa and GPT-2 demonstrated competitive results but lacked the flexibility to adapt to diverse text types, with RoBERTa being more effective on IMDB and GPT-2 showing limitations in bidirectional context modeling. DistilBERT, while computationally efficient, exhibited the lowest performance among the models due to its reduced parameter space. The hybrid model’s superior performance across datasets underscores the value of ensemble learning, which combines the strengths of multiple architectures to overcome their individual limitations.

\subsubsection{Theoretical Implications}
The success of the hybrid model highlights the potential of ensemble methods in natural language processing (NLP). By integrating models with complementary strengths, such as BERT’s contextual modeling and XLNet’s dependency resolution, the hybrid approach addresses challenges that standalone models cannot effectively handle. The findings emphasize that hybrid models can achieve not only high accuracy but also better generalization across diverse datasets, making them suitable for complex NLP tasks beyond sentiment analysis, such as text summarization, question answering, and text generation.

\subsubsection{Enhancements to Computational Efficiency and Deployment Feasibility}
One of the critical considerations in implementing a hybrid transformer-based sentiment analysis framework is its computational efficiency and deployment feasibility. Transformer models such as BERT, GPT-2, XLNet, and RoBERTa are inherently computationally expensive due to their extensive number of parameters and attention-based mechanisms. In our study, we balanced model performance and computational efficiency by integrating DistilBERT, a lightweight transformer model that retains 97\% of BERT’s performance while reducing the number of parameters by 40\%. This significantly lowers the inference time, making the model suitable for real-time applications.

To provide a concrete comparison, we evaluated the training time, inference time, and GPU memory consumption for each model in our hybrid approach. Table \ref{tab:computational_efficiency} presents a comparison between the models in terms of execution speed (seconds per batch), memory footprint, and training time per epoch:
\begin{table}[!t]
\centering
\caption{Computational Efficiency Comparison of Transformer Models in the TWSSenti Framework.}
\label{tab:computational_efficiency}
\scriptsize
\renewcommand{\arraystretch}{1.15}
\setlength{\tabcolsep}{4pt}

\resizebox{\columnwidth}{!}{%
\begin{tabular}{|l|c|c|c|}
\hline
\textbf{Model} &
\makecell{\textbf{Training Time}\\\textbf{per Epoch (min)}} &
\makecell{\textbf{Inference Time}\\\textbf{(ms/batch)}} &
\makecell{\textbf{GPU Memory}\\\textbf{Usage (GB)}} \\ \hline
BERT      & 50 & 120 & 12   \\ \hline
GPT-2     & 75 & 160 & 14   \\ \hline
XLNet     & 65 & 140 & 13   \\ \hline
RoBERTa   & 60 & 130 & 13.5 \\ \hline
DistilBERT& 30 & 80  & 6    \\ \hline
\end{tabular}%
}
\end{table}

As seen in the Table \ref{tab:computational_efficiency}, DistilBERT significantly reduces computation costs without substantially affecting accuracy, making it an optimal component in our hybrid setup.

For deployment feasibility, we explored various cloud-based and edge deployment strategies. Cloud solutions such as Amazon Web Services (AWS), Google Cloud AI, and Microsoft Azure offer scalable inference capabilities through model serving architectures like TensorFlow Serving and ONNX Runtime. Additionally, for real-time applications, techniques such as quantization, pruning, and model distillation can be employed to reduce computational load without sacrificing accuracy. This ensures that the framework can be efficiently deployed on both high-performance GPU clusters and edge devices for tasks such as social media monitoring and opinion tracking.

\subsubsection{Comparative Analysis of Computational Costs and Model Performance}
A crucial aspect of evaluating the effectiveness of our hybrid sentiment analysis framework is understanding how it compares computationally with other state-of-the-art sentiment analysis models. While traditional single-model approaches (e.g., standalone BERT, XLNet, or GPT-2) achieve competitive accuracy, they often require significantly higher computational resources. Our hybrid approach strategically balances multiple transformer models, allowing for superior generalization while optimizing computation costs.

To quantify this, we benchmarked our framework against standard transformer-based models and classical machine learning approaches (e.g., SVM, Random Forest, LSTMs). The results are summarized in Table \ref{tab:comparison_performance}:

\begin{table*}[!t]
\centering
\caption{Comparative Analysis of Computational Cost and Sentiment Classification Performance.}
\label{tab:comparison_performance}
\small
\renewcommand{\arraystretch}{1.05}
\resizebox{\textwidth}{!}{%
\begin{tabular}{|l|c|c|c|c|}
\hline
\textbf{Approach} & \textbf{Accuracy (IMDB)} & \textbf{Accuracy (Sentiment140)} & \textbf{Inference Time (ms)} & \textbf{Computational Cost} \\ \hline
SVM + TF-IDF & 84\% & 82\% & 30  & Low \\ \hline
LSTM         & 89\% & 85\% & 50  & Medium \\ \hline
BERT         & 94\% & 91\% & 120 & High \\ \hline
GPT-2        & 95\% & 93\% & 160 & High \\ \hline
\textbf{Our Hybrid Model} & \textbf{95\%} & \textbf{94\%} & \textbf{100} & \textbf{Medium} \\ \hline
\end{tabular}%
}
\end{table*}

From the Table \ref{tab:comparison_performance}, while GPT-2 achieves high accuracy, it comes at a significant computational cost, making it impractical for real-time inference. Our hybrid approach leverages DistilBERT's efficiency to reduce inference time by 20\% compared to standalone transformer models, while maintaining accuracy comparable to GPT-2.

This computational efficiency makes our model ideal for large-scale sentiment analysis tasks in industry settings where both accuracy and speed are critical.

\subsection{Interpretability and Real-World Applications}
While transformers have achieved impressive accuracy in sentiment analysis, their lack of interpretability remains a critical challenge. The black-box nature of deep learning models often makes it difficult to understand why a specific sentiment classification is made. To improve transparency, we incorporate SHAP (Shapley Additive Explanations) and LIME (Local Interpretable Model-Agnostic Explanations) techniques to explain individual sentiment predictions. These methods help highlight the most influential words or phrases contributing to a sentiment classification, enhancing trust and usability in real-world applications. Additionally, attention weight visualization from transformer models enables further insight into how contextual dependencies shape sentiment predictions, making it easier to analyze sentiment trends across different domains.

Beyond social media monitoring, our framework offers practical applications in financial sentiment analysis, customer feedback evaluation, and brand perception tracking. For instance, analyzing sentiment in financial news and investor discussions can provide valuable insights into market trends. Similarly, in e-commerce and customer review analysis, identifying sentiment shifts enables businesses to adapt marketing strategies and product offerings. Moreover, the ability to monitor political sentiment and public opinion can support decision-making in governance and policymaking. These real-world applications highlight the scalability and adaptability of our hybrid sentiment analysis framework across various domains.

\subsection{Quantifying Performance Improvements}
To provide a clearer performance comparison, we explicitly state the percentage improvement of our model over traditional approaches. Our hybrid framework achieves 94\% accuracy on Sentiment140 and 95\% on IMDB, representing a 5\% increase over standalone BERT and a 30\% reduction in computational cost compared to GPT-2. These results highlight the efficiency gains and performance advantages conferred by our model fusion strategy. The revised abstract now includes these quantitative improvements, ensuring clarity in the manuscript's key contributions.

\subsubsection{Practical Applications}
The proposed hybrid model has significant real-world implications. Its ability to handle noisy, informal social media data makes it highly suitable for applications such as social media monitoring, where businesses and organizations can analyze public sentiment toward products, services, or events. The model’s robustness on long-form text suggests its applicability in domains such as movie review analysis, customer feedback assessment, and e-commerce review monitoring. Furthermore, the model’s scalability enables its deployment in real-time sentiment analysis systems, empowering businesses with actionable insights for decision-making. For example, marketing teams can use the model to gauge public response to advertising campaigns, while political analysts can monitor public sentiment during elections or major events.

\subsubsection{Limitations}
Despite its strengths, the hybrid model is not without limitations. One significant challenge is the computational overhead associated with training and fine-tuning multiple transformer-based models. Additionally, the model’s performance, though superior, may still be affected by highly imbalanced datasets, where one sentiment class dominates the others. Another limitation lies in handling domain-specific language or emerging slang that may not be well-represented in the pre-trained embeddings of BERT and XLNet.

\subsubsection{Future Directions}
Several avenues for future research can build on the findings of this study. First, optimizing the hybrid model for efficiency, such as incorporating lightweight transformer architectures like DistilBERT or TinyBERT, could reduce computational requirements without compromising performance. Second, expanding the evaluation to include more diverse datasets, such as those involving multilingual text or specific domains like healthcare or finance, would enhance the model’s generalizability. Third, exploring advanced data augmentation techniques to address class imbalance and improve performance on underrepresented sentiment classes is a promising direction. Finally, integrating sentiment-aware embeddings or external knowledge graphs into the hybrid model could further enhance its ability to capture domain-specific nuances and contextual information.

\subsection{Comparison to the State-of-the-Art Methods} \label{comparison_sota:graphical} 

In this subsection, we demonstrate the effectiveness of our hybrid sentiment analysis framework by comparing it with existing state-of-the-art methods evaluated on the IMDB and Sentiment140 datasets. Table~\ref{tab:comparison_sota} presents a comparative analysis of various models based on their reported testing accuracies.

Our proposed hybrid approach, which integrates BERT, GPT-2, RoBERTa, XLNet, and DistilBERT, achieves superior performance with an accuracy of $95.00\%$ on IMDB and $94.00\%$ on Sentiment140. This surpasses the best-performing alternative model, RoBERTa-GRU by ~\cite{tan2023roberta}, which achieves $94.63\%$ on IMDB ($+0.37\%$ improvement) and $89.59\%$ on Sentiment140 ($+4.41\%$ improvement). The significant increase in accuracy on Sentiment140 demonstrates the robustness of our model in handling noisy social media text. Similarly, ~\cite{rahman2024roberta} proposed a RoBERTa-BiLSTM model, which achieved $92.36\%$ on IMDB ($+2.64\%$ improvement) and $82.25\%$ on Sentiment140 ($+11.75\%$ improvement). While this model provides a competitive performance, its lower accuracy on Sentiment140 suggests limitations in generalizing across diverse textual sources.

~\cite{bouassida2025enhancing} experimented with multiple hybrid transformer models, including DistilBERT-BiLSTM, BERT-BiLSTM, and RoBERTa-CNN-BiLSTM, achieving maximum testing accuracies of $81.00\%$, $79.00\%$, and $77.00\%$ on Sentiment140. Our model significantly outperforms these approaches, with an improvement of up to $+17.00\%$ over DistilBERT-BiLSTM, indicating that a deeper fusion of multiple transformers provides enhanced feature extraction and sentiment classification. ~\cite{rashidi2025sssa} introduced a semi-supervised boosting framework (SSSA), obtaining $79.30\%$ on IMDB and $80.30\%$ on Sentiment140. While their approach benefits from semi-supervised learning, its performance remains substantially lower than our hybrid method, with a $+15.70\%$ increase on IMDB and a $+13.70\%$ increase on Sentiment140. This highlights the advantage of using fully supervised transformer ensembles for large-scale sentiment classification.

Although some existing models, such as RoBERTa-GRU, perform competitively, they often come with increased computational overhead due to complex architectures and larger parameter sizes. In contrast, our hybrid approach optimally balances performance and computational efficiency by leveraging multiple transformers selectively during training and inference. The observed performance gains suggest that integrating diverse transformer-based representations enhances sentiment prediction accuracy while maintaining practical deployment feasibility.

\begin{table*}[!t]
\centering
\caption{Comparative Analysis of Sentiment Classification Performance Across Models.}
\label{tab:comparison_sota}
\small
\renewcommand{\arraystretch}{1.05}
\resizebox{\textwidth}{!}{%
\begin{tabular}{|c|p{5.3cm}|p{3.0cm}|c|}
\hline
\textbf{Ref.} & \textbf{Model} & \textbf{Dataset} & \textbf{Testing Accuracy (\%)} \\ \hline
{[41]} & RoBERTa-GRU & IMDB, Sentiment140 & 94.63, 89.59 \\ \hline
{[30]} & RoBERTa-BiLSTM & IMDB, Sentiment140 & 92.36, 82.25 \\ \hline
{[9]}  & DistilBERT-BiLSTM, BERT-BiLSTM, RoBERTa-CNN-BiLSTM & Sentiment140 & 81, 79, 77 \\ \hline
{[33]} & SSSA (Semi-Supervised Boosting) & IMDB, Sentiment140 & 79.3, 80.3 \\ \hline
\textbf{Ours} & \textbf{Hybrid (BERT, GPT-2, RoBERTa, XLNet, DistilBERT)} & \textbf{IMDB, Sentiment140} & \textbf{95.00, 94.00} \\ \hline
\end{tabular}%
}
\end{table*}

\section{Conclusion and Future Works} \label{conclusion:graphical}
This study proposed a novel hybrid model for sentiment analysis, integrating BERT, GPT-2, XLNet, RoBERTa, and DistilBERT to leverage the complementary strengths of these models. The hybrid model demonstrated superior performance on both the Sentiment140 and IMDB datasets, achieving higher accuracy, precision, recall, and F1-scores compared to standalone transformer-based models. By combining BERT's bidirectional context understanding, XLNet's permutation-based language modeling, and CatBoost's ability to aggregate predictions and mitigate overfitting, the hybrid model effectively addressed challenges such as noisy text, long-form reviews, and nuanced sentiment classification. The comprehensive evaluation highlighted its robustness and scalability, making it suitable for diverse sentiment analysis tasks across social media and review platforms. The preprocessing pipeline, including text cleaning, tokenization, stopword removal, and feature extraction using TF-IDF, played a critical role in enhancing model performance. These steps ensured high-quality inputs, enabling the model to focus on meaningful patterns within the text. The hybrid model also successfully mitigated classification errors, as evidenced by the confusion matrix analysis, which revealed significant reductions in false positives and false negatives across sentiment classes. Despite its strengths, the hybrid model faces certain limitations, such as high computational requirements for training and sensitivity to class imbalances in the datasets. These limitations highlight the need for further optimization to improve efficiency and adaptability for real-world deployment.

Future research can build upon this study in several directions. First, optimizing the hybrid model by incorporating lightweight architectures like DistilBERT or TinyBERT can reduce computational overhead while maintaining competitive performance. This would make the model more accessible for researchers and organizations with limited hardware resources. Second, expanding the evaluation to include additional datasets, such as multilingual corpora or domain-specific text in healthcare, finance, or legal domains, could enhance the model’s applicability and generalizability across diverse use cases. Addressing class imbalances in datasets is another promising avenue for future work. Techniques such as data augmentation, synthetic oversampling, or cost-sensitive learning can be explored to improve the model's performance on underrepresented sentiment classes. Additionally, integrating external knowledge sources, such as knowledge graphs or sentiment-aware embeddings, could further enhance the model's contextual understanding, particularly for domain-specific or emerging language trends. Finally, exploring the hybrid model's application to related NLP tasks, such as emotion detection, sarcasm detection, and opinion summarization, would broaden its utility. Developing real-time sentiment analysis systems based on this hybrid framework could empower businesses, researchers, and policymakers with actionable insights, facilitating decision-making in dynamic environments.

\nocite{*}
\bibliographystyle{plain}
\bibliography{references}

\begin{thebibliography}{10}

\bibitem{abedin2022bangla}
Md~Min-ha-zul Abedin, Tapotosh Ghosh, Tazqia Mehrub, and Mohammad~Abu Yousuf.
\newblock Bangla printed character generation from handwritten character using gan.
\newblock In {\em Soft computing for data analytics, classification model, and control}, pages 153--165. Springer, 2022.

\bibitem{abimbola2024enhancing}
Bolanle Abimbola, Enrique de~La~Cal~Marin, and Qing Tan.
\newblock Enhancing legal sentiment analysis: A convolutional neural network--long short-term memory document-level model.
\newblock {\em Machine Learning and Knowledge Extraction}, 6(2):877--897, 2024.

\bibitem{al2024systematic}
Hussein Farooq~Tayeb Al-Saadawi, Bihter Das, and Resul Das.
\newblock A systematic review of trimodal affective computing approaches: Text, audio, and visual integration in emotion recognition and sentiment analysis.
\newblock {\em Expert Systems with Applications}, page 124852, 2024.

\bibitem{alfreihat2024emo}
Manar Alfreihat, Omar Almousa, Yahya Tashtoush, Anas AlSobeh, Khalid Mansour, and Hazem Migdady.
\newblock Emo-sl framework: Emoji sentiment lexicon using text-based features and machine learning for sentiment analysis.
\newblock {\em IEEE Access}, 2024.

\bibitem{aljrees2024contradiction}
Turki Aljrees, Muhammad Umer, Oumaima Saidani, Latifah Almuqren, Abid Ishaq, Shtwai Alsubai, Imran Ashraf, et~al.
\newblock Contradiction in text review and apps rating: prediction using textual features and transfer learning.
\newblock {\em PeerJ Computer Science}, 10:e1722, 2024.

\bibitem{anderson2024analyzing}
Tess Anderson, Sayani Sarkar, and Robert Kelley.
\newblock Analyzing public sentiment on sustainability: A comprehensive review and application of sentiment analysis techniques.
\newblock {\em Natural Language Processing Journal}, page 100097, 2024.

\bibitem{ashayeri2024unraveling}
Mehdi Ashayeri and Narjes Abbasabadi.
\newblock Unraveling energy justice in nyc urban buildings through social media sentiment analysis and transformer deep learning.
\newblock {\em Energy and Buildings}, 306:113914, 2024.

\bibitem{barbierato2024challenges}
Enrico Barbierato and Alice Gatti.
\newblock The challenges of machine learning: A critical review.
\newblock {\em Electronics}, 13(2):416, 2024.

\bibitem{bouassida2025enhancing}
Y~Bouassida and H~Mezali.
\newblock Enhancing twitter sentiment analysis using hybrid transformer and sequence models.
\newblock {\em Japan J Res}, 6(1):089, 2025.

\bibitem{boutsikaris2024comparative}
Leonidas Boutsikaris and Spyros Polykalas.
\newblock A comparative review of deep learning techniques on the classification of irony and sarcasm in text.
\newblock {\em IEEE Transactions on Artificial Intelligence}, 2024.

\bibitem{catelli2022lexicon}
Rosario Catelli, Serena Pelosi, and Massimo Esposito.
\newblock Lexicon-based vs. bert-based sentiment analysis: A comparative study in italian.
\newblock {\em Electronics}, 11(3):374, 2022.

\bibitem{chang2014understanding}
Ray~M Chang, Robert~J Kauffman, and YoungOk Kwon.
\newblock Understanding the paradigm shift to computational social science in the presence of big data.
\newblock {\em Decision support systems}, 63:67--80, 2014.

\bibitem{chaudhary2023review}
Laxmi Chaudhary, Nancy Girdhar, Deepak Sharma, Javier Andreu-Perez, Antoine Doucet, and Matthias Renz.
\newblock A review of deep learning models for twitter sentiment analysis: Challenges and opportunities.
\newblock {\em IEEE Transactions on Computational Social Systems}, 2023.

\bibitem{devika2024book}
P~Devika and A~Milton.
\newblock Book recommendation using sentiment analysis and ensembling hybrid deep learning models.
\newblock {\em Knowledge and Information Systems}, pages 1--38, 2024.

\bibitem{dey2024machine}
Arindam Dey, Sukanta Nayak, Ranjan Kumar, and Sachi~Nandan Mohanty.
\newblock {\em How Machine Learning is Innovating Today's World: A Concise Technical Guide}.
\newblock John Wiley \& Sons, 2024.

\bibitem{gandhi2023multimodal}
Ankita Gandhi, Kinjal Adhvaryu, Soujanya Poria, Erik Cambria, and Amir Hussain.
\newblock Multimodal sentiment analysis: A systematic review of history, datasets, multimodal fusion methods, applications, challenges and future directions.
\newblock {\em Information Fusion}, 91:424--444, 2023.

\bibitem{hu2014toward}
Han Hu, Yonggang Wen, Tat-Seng Chua, and Xuelong Li.
\newblock Toward scalable systems for big data analytics: A technology tutorial.
\newblock {\em IEEE access}, 2:652--687, 2014.

\bibitem{jiang2024dcasam}
Xiangkui Jiang, Binglong Ren, Qing Wu, Wuwei Wang, and Hong Li.
\newblock Dcasam: advancing aspect-based sentiment analysis through a deep context-aware sentiment analysis model.
\newblock {\em Complex \& Intelligent Systems}, 10(6):7907--7926, 2024.

\bibitem{jim2024recent}
Jamin~Rahman Jim, Md~Apon~Riaz Talukder, Partha Malakar, Md~Mohsin Kabir, Kamruddin Nur, and MF~Mridha.
\newblock Recent advancements and challenges of nlp-based sentiment analysis: A state-of-the-art review.
\newblock {\em Natural Language Processing Journal}, page 100059, 2024.

\bibitem{jin2024wordtransabsa}
Weiqiang Jin, Biao Zhao, Yu~Zhang, Jia Huang, and Hang Yu.
\newblock Wordtransabsa: enhancing aspect-based sentiment analysis with masked language modeling for affective token prediction.
\newblock {\em Expert Systems with Applications}, 238:122289, 2024.

\bibitem{li2022review}
Ruiguang Li, M~Liu, D~Xu, J~Gao, and F~Wu.
\newblock A review of machine learning algorithms for text classification.
\newblock {\em Cyber Security}, 226, 2022.

\bibitem{lin2024deep}
Zhe Lin, Jiwei Tan, Dan Ou, Xi~Chen, Shaowei Yao, and Bo~Zheng.
\newblock Deep bag-of-words model: An efficient and interpretable relevance architecture for chinese e-commerce.
\newblock In {\em Proceedings of the 30th ACM SIGKDD Conference on Knowledge Discovery and Data Mining}, pages 5398--5408, 2024.

\bibitem{lukwaro2024review}
Elia Ahidi~Elisante Lukwaro, Khamisi Kalegele, and Devotha~G Nyambo.
\newblock A review on nlp techniques and associated challenges in extracting features from education data.
\newblock {\em Int. J. Com. Dig. Sys}, 16(1), 2024.

\bibitem{mahadevkar2024exploring}
Supriya~V Mahadevkar, Shruti Patil, Ketan Kotecha, Lim~Way Soong, and Tanupriya Choudhury.
\newblock Exploring ai-driven approaches for unstructured document analysis and future horizons.
\newblock {\em Journal of Big Data}, 11(1):92, 2024.

\bibitem{manasa2024detection}
Pinnapureddy Manasa, Arun Malik, and Isha Batra.
\newblock Detection of twitter spam using glove vocabulary features, bidirectional lstm and convolution neural network.
\newblock {\em SN Computer Science}, 5(2):206, 2024.

\bibitem{min2023stegopix2pix}
Md~Min-ha-zul Abedin and Mohammad~Abu Yousuf.
\newblock Stegopix2pix: Image steganography method via pix2pix networks.
\newblock In {\em Proceedings of the Fourth International Conference on Trends in Computational and Cognitive Engineering: TCCE 2022}, pages 343--356. Springer, 2023.

\bibitem{naithani2023realization}
Kanchan Naithani and Yadav~Prasad Raiwani.
\newblock Realization of natural language processing and machine learning approaches for text-based sentiment analysis.
\newblock {\em Expert Systems}, 40(5):e13114, 2023.

\bibitem{oralbekova2023contemporary}
Dina Oralbekova, Orken Mamyrbayev, Mohamed Othman, Dinara Kassymova, and Kuralai Mukhsina.
\newblock Contemporary approaches in evolving language models.
\newblock {\em Applied Sciences}, 13(23):12901, 2023.

\bibitem{perera2021big}
Ananda Perera and Khurshed Iqbal.
\newblock Big data and emerging markets: Transforming economies through data-driven innovation and market dynamics.
\newblock {\em Journal of Computational Social Dynamics}, 6(3):1--18, 2021.

\bibitem{rahman2024roberta}
Md~Mostafizer Rahman, Ariful~Islam Shiplu, Yutaka Watanobe, and Md~Ashad Alam.
\newblock Roberta-bilstm: A context-aware hybrid model for sentiment analysis.
\newblock {\em arXiv preprint arXiv:2406.00367}, 2024.

\bibitem{raiaan2024review}
Mohaimenul Azam~Khan Raiaan, Md~Saddam~Hossain Mukta, Kaniz Fatema, Nur~Mohammad Fahad, Sadman Sakib, Most Marufatul~Jannat Mim, Jubaer Ahmad, Mohammed~Eunus Ali, and Sami Azam.
\newblock A review on large language models: Architectures, applications, taxonomies, open issues and challenges.
\newblock {\em IEEE Access}, 2024.

\bibitem{raines2024enhancing}
Hayden Raines, Eduardo Ferreira, Lysander Fitzwilliam, Balthazar Everson, Raphael Petrenko, and Constantine Grimaldi.
\newblock Enhancing large language models with stochastic multi-level embedding fusion: An experimental approach on open-source llm.
\newblock {\em Authorea Preprints}, 2024.

\bibitem{rashidi2025sssa}
Shima Rashidi, Jafar Tanha, Arash Sharifi, and Mehdi Hosseinzadeh.
\newblock Sssa: low data sentiment analysis using boosting semi-supervised approach and deep feature learning network.
\newblock {\em Applied Intelligence}, 55(4):313, 2025.

\bibitem{sham2022climate}
Nabila~Mohamad Sham and Azlinah Mohamed.
\newblock Climate change sentiment analysis using lexicon, machine learning and hybrid approaches.
\newblock {\em Sustainability}, 14(8):4723, 2022.

\bibitem{sharma2024review}
Neeraj~Anand Sharma, ABM~Shawkat Ali, and Muhammad~Ashad Kabir.
\newblock A review of sentiment analysis: tasks, applications, and deep learning techniques.
\newblock {\em International Journal of Data Science and Analytics}, pages 1--38, 2024.

\bibitem{shayaa2018sentiment}
Shahid Shayaa, Noor~Ismawati Jaafar, Shamshul Bahri, Ainin Sulaiman, Phoong~Seuk Wai, Yeong~Wai Chung, Arsalan~Zahid Piprani, and Mohammed~Ali Al-Garadi.
\newblock Sentiment analysis of big data: methods, applications, and open challenges.
\newblock {\em Ieee Access}, 6:37807--37827, 2018.

\bibitem{shuqin2024deep}
Huang Shuqin and Rodolfo~C Raga.
\newblock A deep learning model for student sentiment analysis on course reviews.
\newblock {\em IEEE Access}, 2024.

\bibitem{srivastava2022comparative}
Roopam Srivastava, PK~Bharti, and Parul Verma.
\newblock Comparative analysis of lexicon and machine learning approach for sentiment analysis.
\newblock {\em International Journal of Advanced Computer Science and Applications}, 13(3):71--77, 2022.

\bibitem{sufi2024sustainable}
Fahim Sufi.
\newblock A sustainable way forward: Systematic review of transformer technology in social-media-based disaster analytics.
\newblock {\em Sustainability}, 16(7):2742, 2024.

\bibitem{sundaram2023systematic}
Aishwarya Sundaram, Hema Subramaniam, Siti~Hafizah Ab~Hamid, and Azmawaty~Mohamad Nor.
\newblock A systematic literature review on social media slang analytics in contemporary discourse.
\newblock {\em IEEE Access}, 11:132457--132471, 2023.

\bibitem{tan2023roberta}
Kian~Long Tan, Chin~Poo Lee, and Kian~Ming Lim.
\newblock Roberta-gru: A hybrid deep learning model for enhanced sentiment analysis.
\newblock {\em Applied Sciences}, 13(6):3915, 2023.

\bibitem{tan2023survey}
Kian~Long Tan, Chin~Poo Lee, and Kian~Ming Lim.
\newblock A survey of sentiment analysis: Approaches, datasets, and future research.
\newblock {\em Applied Sciences}, 13(7):4550, 2023.

\bibitem{uddin2025spd}
Kaosar Uddin.
\newblock spd-metrics-id: A python package for spd-aware distance metrics in connectome fingerprinting and beyond.
\newblock {\em arXiv preprint arXiv:2510.04438}, 2025.

\bibitem{uddin2020comparative}
Kutub Uddin, Md~Kaosar Uddin, Farhad Kadir, and Rabindra~Nath Mondal.
\newblock A comparative study on different types of premiums in life insurance policies.
\newblock {\em European Journal of Business and Management Research}, 5(6), 2020.

\bibitem{uddin2025alpha}
Md~Kaosar Uddin, Nghi Nguyen, Huajun Huang, Duy Duong-Tran, and Jingyi Zheng.
\newblock Alpha-z divergence unveils further distinct phenotypic traits of human brain connectivity fingerprint.
\newblock {\em arXiv preprint arXiv:2507.23116}, 2025.

\bibitem{uddin2026divergence}
Md~Kaosar Uddin, Nghi Nguyen, Huajun Huang, Duy Duong-Tran, and Jingyi Zheng.
\newblock Divergence unveils further distinct phenotypic traits of human brain connectomics fingerprint.
\newblock {\em iScience}, 29(1), 2026.

\bibitem{viskovatykh2023exploring}
Evgeniya Viskovatykh et~al.
\newblock Exploring figurative language recognition: a comprehensive study of human and machine approaches.
\newblock 2023.

\bibitem{wankhade2024cbmafm}
Mayur Wankhade, Chandra Sekhara~Rao Annavarapu, and Ajith Abraham.
\newblock Cbmafm: Cnn-bilstm multi-attention fusion mechanism for sentiment classification.
\newblock {\em Multimedia Tools and Applications}, 83(17):51755--51786, 2024.

\bibitem{wankhade2022survey}
Mayur Wankhade, Annavarapu Chandra~Sekhara Rao, and Chaitanya Kulkarni.
\newblock A survey on sentiment analysis methods, applications, and challenges.
\newblock {\em Artificial Intelligence Review}, 55(7):5731--5780, 2022.

\bibitem{williams2023natural}
Samantha Williams and Elena Petrovich.
\newblock Natural language processing for unlocking insights from unstructured big data in the healthcare industry.
\newblock {\em Emerging Trends in Machine Intelligence and Big Data}, 15(10):30--39, 2023.

\bibitem{xu2022systematic}
Qianwen~Ariel Xu, Victor Chang, and Chrisina Jayne.
\newblock A systematic review of social media-based sentiment analysis: Emerging trends and challenges.
\newblock {\em Decision Analytics Journal}, 3:100073, 2022.

\bibitem{zhang2014machine}
Hailong Zhang, Wenyan Gan, and Bo~Jiang.
\newblock Machine learning and lexicon based methods for sentiment classification: A survey.
\newblock In {\em 2014 11th web information system and application conference}, pages 262--265. IEEE, 2014.

\bibitem{zhou2024comprehensive}
Ce~Zhou, Qian Li, Chen Li, Jun Yu, Yixin Liu, Guangjing Wang, Kai Zhang, Cheng Ji, Qiben Yan, Lifang He, et~al.
\newblock A comprehensive survey on pretrained foundation models: A history from bert to chatgpt.
\newblock {\em International Journal of Machine Learning and Cybernetics}, pages 1--65, 2024.

\bibitem{zimbra2018state}
David Zimbra, Ahmed Abbasi, Daniel Zeng, and Hsinchun Chen.
\newblock The state-of-the-art in twitter sentiment analysis: A review and benchmark evaluation.
\newblock {\em ACM Transactions on Management Information Systems (TMIS)}, 9(2):1--29, 2018.

\bibitem{zong2024ai}
Zhijuan Zong and Yu~Guan.
\newblock Ai-driven intelligent data analytics and predictive analysis in industry 4.0: Transforming knowledge, innovation, and efficiency.
\newblock {\em Journal of the Knowledge Economy}, pages 1--40, 2024.

\end{thebibliography}

\end{document}